%% file: 0main.tex
\documentclass{sigkddExp}
\usepackage{etoolbox}
\makeatletter
\patchcmd{\@sect}{\vskip -14pt}{\vskip 0pt}{}{}
\makeatother

\usepackage{amsmath}
\usepackage{mathtools}
\usepackage{amsthm}
\usepackage{graphicx}
\usepackage{booktabs}
\usepackage{subcaption}
\usepackage{adjustbox}
\usepackage{url}
\usepackage{wrapfig}
\usepackage{hyperref}
\usepackage{soul}
\usepackage{multirow}
\usepackage[textsize=tiny]{todonotes}
\usepackage{algorithm}
\usepackage{algorithmic}
\usepackage[numbers]{natbib} 
\makeatletter
\renewcommand\@biblabel[1]{[#1]}
\usepackage{bbm}
\usepackage{listings}
\usepackage{xspace}

\usepackage{multirow}
\usepackage{multicol}

\newtheorem{definition}{Definition}

\newcommand{\methodFont}{\texttt}
\newcommand{\ours}{\methodFont{SEP}\xspace}

\newcommand{\expl}{\mathit{I}_f\xspace}

\newcommand{\metricFont}{\textit}
\newcommand{\accAdv}{\ensuremath{\metricFont{Acc}_{\text{adv}}}\xspace}
\newcommand{\accClean}{\ensuremath{\metricFont{Acc}_{\text{clean}}}\xspace}

\usepackage{xcolor,colortbl}

\definecolor{Gray}{gray}{0.85}
\definecolor{LightCyan}{rgb}{0.88,1,1}

\newcolumntype{a}{>{\columncolor{Gray}}c}
\newcolumntype{b}{>{\columncolor{white}}c}

\usepackage[capitalize,noabbrev]{cleveref}

\providecommand{\argmin}{\mathop{\mathrm{arg\,min}}}

\begin{document}
%

\title{Are Classification Robustness and Explanation Robustness Really Strongly Correlated? An Analysis Through Input Loss Landscape}
%

\numberofauthors{3}
%


\author{
%
\alignauthor Tiejin Chen \\
       \affaddr{Arizona State University}\\
       \email{ tchen169@asu.edu}
\alignauthor Wenwang Huang\\
       \affaddr{Independent Researcher}\\
    \email{Weistrasshww@gmail.com}
\and
\alignauthor Linsey Pang \\
       \affaddr{Paypal AI}\\
       \email{panglinsey@gmail.com}
\alignauthor Dongsheng Luo \\
       \affaddr{Florida International University}\\
       \email{dluo@fiu.edu}
\alignauthor Hua Wei \\
       \affaddr{Arizona State University}\\
       \email{hua.wei@asu.edu}
}

\maketitle
\begin{abstract}
This paper looks into the critical area of deep learning robustness and challenges the common belief that classification robustness and explanation robustness in image classification systems are inherently correlated.  Through a novel evaluation approach leveraging clustering for efficient assessment of explanation robustness, we demonstrate that enhancing explanation robustness does not necessarily flatten the input loss landscape with respect to explanation loss - contrary to flattened loss landscapes indicating better classification robustness. To further investigate this contradiction, a training method designed to adjust the loss landscape with respect to explanation loss is proposed. Through the new training method, we uncover that although such adjustments can impact the robustness of explanations, they do not have an influence on the robustness of classification. These findings not only challenge the previous assumption of a strong correlation between the two forms of robustness but also pave new pathways for understanding the relationship between loss landscape and explanation loss. Codes are provided in \url{https://github.com/tiejin98/relationship_two_robustness}.

\end{abstract}

\input{1intro_hua}
\input{1related_hua}
\input{2method_hua}

\input{3experiment}
\input{4conclusion}

\bibliographystyle{abbrv}
\bibliography{sigproc}  
\clearpage
\input{5appendix}

\end{document}

%% file: 1intro_hua.tex
\section{Introduction}

Understanding the relationship between classification robustness and explanation robustness is critical for deploying reliable machine learning systems in real-world applications. In medical diagnostics, autonomous vehicles, and financial fraud detection, models must not only maintain accurate predictions under adversarial attacks (known as classification robustness~\citep{madry2017towards}) but also preserve consistent, interpretable rationales (e.g., highlighted image regions) for their decisions during adversarial attacks (known as explanation robustness). Prior work has widely assumed these two properties are positively correlated~\citep{boopathy2020proper,huang2023safari}, leading to potential security gaps if they are not directly linked.

\begin{figure}
     \centering
     \includegraphics[width=1.0\linewidth]{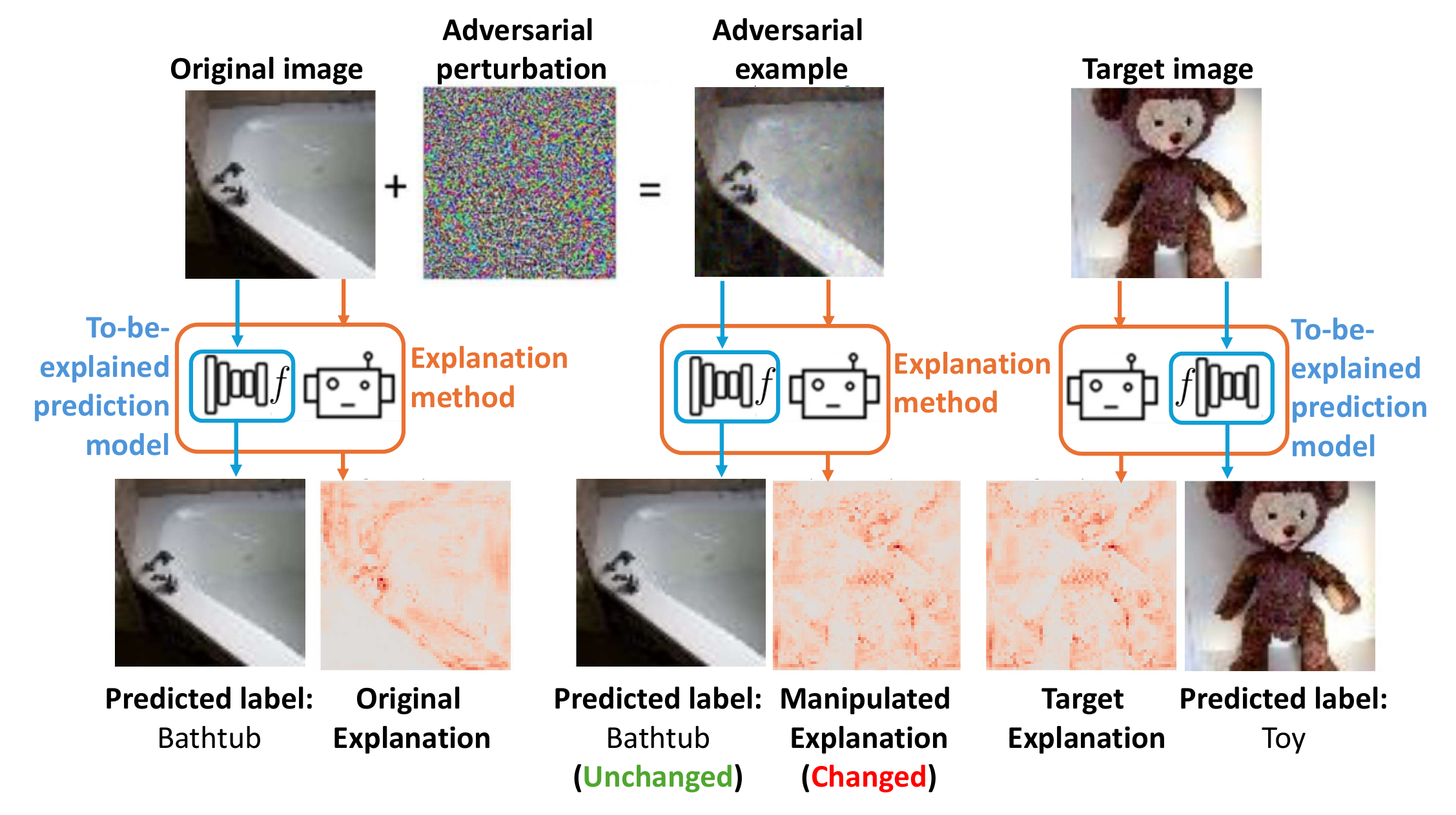}

\caption{Example of an adversarial explanation attack. While the to-be-explained model $f$ gives the same predicted label after adding adversarial noise, the explanation saliency map can be manipulated to the target saliency map, leading to wrong explanations.}
\label{fig:attack}
\end{figure}

Our investigation reveals a fundamental challenge to this paradigm: \textit{improving classification robustness provides no guarantee of enhanced explanation robustness}, challenging a key assumption in adversarial learning. This occurs because explanations, such as saliency maps, are themselves susceptible to adversarial perturbations~\citep{dombrowski2019explanations,ghorbani2019interpretation}. As illustrated in~\cref{fig:attack}, small adversarial perturbations can significantly alter explanation maps while leaving the model’s classification unchanged. This raises a critical question about whether adversarial robustness in classification translates to robustness in model interpretability.

\begin{figure}
\includegraphics[width=0.75\linewidth]{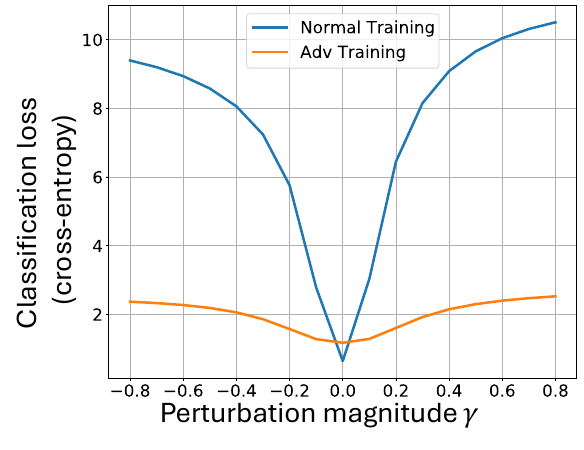}
\caption{Comparison of input loss landscapes between normal and adversarially trained models on CIFAR-10. Adversarially trained models show flatter loss landscapes and increased classification robustness.}
\label{fig:input_ce}
\end{figure}

To better understand robustness, one key approach is analyzing the input loss landscape~\citep{li2023understanding}. Prior work has shown that a flatter input loss landscape with respect to classification loss indicates stronger classification robustness~\citep{xie2020smooth,li2023understanding}. As visualized in~\cref{fig:input_ce}, adversarially trained models, which show increased classification robustness, exhibit a flatter input loss landscape compared to normally trained models. Since classification robustness is linked to a flatter loss landscape, a natural question arises: 
    \textit{Does increasing explanation robustness lead to a flatter input loss landscape for explanation loss?}

Surprisingly, this paper shows that increasing explanation robustness does not flatten the input loss landscape with respect to explanation loss. To systematically analyze the relationship between explanation robustness and input loss landscapes, we generate models with varying levels of explanation robustness using adversarial training methods~\citep{zhang2019theoretically} that allow fine-grained control over classification robustness. Our findings (detailed in~\cref{fig:flat_Trade}) contradict previous assumptions that classification and explanation robustness are inherently linked.

To further investigate this phenomenon, we reverse the question: 
    \textit{Does a flatter input loss landscape for explanation loss improve explanation robustness?} 
This paper shows that the answer is no. Flattening the input loss landscape for explanation loss actually decreases explanation robustness. To confirm this, we propose \textbf{S}eparate \textbf{E}xplanation Robustness via \textbf{P}GD (\ours) to explicitly control the input loss landscape w.r.t explanation robustness. By regularizing on the input loss landscape w.r.t explanation robustness, \ours reduces explanation robustness while leaving classification robustness unchanged. 
Extensive results on different model architectures, datasets, and explanation methods demonstrate the effectiveness of \ours in decoupling explanation robustness and classification robustness. This demonstration challenges the widely held belief that classification and explanation robustness are positively correlated. In summary, the key contributions of this paper are as follows:
\begin{itemize}
    \item We introduce a clustering-based sampling method to efficiently evaluate explanation robustness.
    \item We use TRADES~\citep{zhang2019theoretically} to control classification robustness and visualize the input loss landscape with respect to explanation loss, revealing that increasing explanation robustness does not flatten the input loss landscape, which contradicts with classification robustness.
    \item We develop a novel training method that flattens the input loss landscape for explanation loss and show that, contrary to prior assumptions, this reduces explanation robustness, suggesting that explanation and classification robustness are not strongly correlated.
\end{itemize}

%% file: 1related_hua.tex
\section{Related Work}
\textbf{Adversarial Attack and Adversarial Training (AT).} It has been proven that deep learning models are vulnerable to adversarial examples~\citep{szegedy2013intriguing,goodfellow2014explaining,carlini2017towards}, where noise that is imperceptible to humans, when added to the original inputs, can lead to the misclassification of models. Projected Gradient Descent (PGD)~\citep{madry2017towards} is one of the most popular methods that generate such a noise or evaluate models' classification robustness by calculating accuracy under its attack. Many methods~\citep{papernot2016distillation,xu2017feature,lin2019defensive,song2017pixeldefend,carlini2022certified} have been introduced to defend against adversarial attacks, while they do not involve a training process and may be vulnerable to adaptive attack~\citep{athalye2018obfuscated}. Goodfellow et al.~\citep{goodfellow2014explaining} first introduced adversarial training (AT), which trains a model from scratch with adversarial samples and proves its performance, including adversarial competitions~\citep{shafahi2019adversarial,liu2021using,madry2017towards,brendel2020adversarial}. In this paper, we focus on classification robustness increased by AT methods like Madry adversarial training~\citep{madry2017towards} and TRADES~\citep{zhang2019theoretically}. 
Many works tend to increase the performance of AT through external datasets~\citep{hendrycks2019using,carmon2019unlabeled,wang2023better}, metric learning~\citep{pang2019rethinking}, self-supervised learning~\citep{chen2020self}, ensemble learning~\citep{tramer2017ensemble}, label smoothing~\citep{chen2020robust}, and Taylor Expansion~\citep{jin2023randomized}. Wu et al.~\citep{wu2020adversarial} found that obtaining a flat loss landscape can help increase classification robustness, which inspired the ideas in this paper. 


\vspace{1mm}
\noindent\textbf{Explanation Robustness.}
Saliency maps~\cite{simonyan2013deep,shrikumar2017learning,bach2015pixel,selvaraju2016grad} are widely used to explain image-related tasks in deep learning, and our focus is on the robustness of these explanations. However, similar to an adversarial attack, it is possible to find an adversarial noise on original images so that it can easily manipulate the saliency maps without changing classification results in both white-box~\cite{dombrowski2019explanations,ghorbani2019interpretation,heo2019fooling,slack2020fooling} and black-box settings~\cite{tamam2022foiling}. Zhang et al.~\cite{zhang2020interpretable} further introduced a new method that can attack both saliency maps and classification results. In order to evaluate the explanation robustness, Wicker et al.~\cite{wicker2022robust} introduced the max-sensitivity and average-sensitivity of saliency maps. Alvarez et al.~\cite{alvarez2018robustness} estimated explanation robustness by the Local Lipschitz of interpretation, while Tamam et al.~\cite{tamam2022foiling} directly used attack loss to evaluate explanation robustness. In this paper, we use attack loss based on the proposed cluster method to evaluate explanation robustness.

Several works have also aimed to improve explanation robustness. Chen et al.~\cite{chen2019robust} introduced a regularization term during training to make the explanation more robust. Boopathy et al.~\cite{boopathy2020proper} improved the performance by training with noisy labels. Tang et al.~\cite{tang2022defense} proposed a first-order gradient-based approach to reduce computational training costs. Huang et al.~\cite{huang2023safari} explored genetic algorithms to optimize for stronger explanation robustness.

\vspace{1mm}
\noindent\textbf{Relationship between Classification Robustness and Explanation.} It has been suggested that improved interpretability contributes to classification robustness~\citep{simonyan2019deep,wang2021robust}, implying a correlation between high-quality saliency maps and classification performance. Studies have demonstrated that models robust to explanation attacks tend to be more resistant to classification attacks as well, supporting the idea that explanation robustness benefits classification robustness~\citep{boopathy2020proper,tang2022defense,huang2023safari}.
However, our work challenges this assumption. We show that adversarial training with TRADES~\citep{zhang2019theoretically} can improve explanation robustness while sometimes enhancing classification robustness. Yet, further analysis reveals that these two aspects of robustness are not inherently linked—classification robustness does not necessarily ensure explanation robustness. This finding suggests a fundamental disconnect between the two, which we investigate in depth.

%% file: 2method_hua.tex
\section{Analysis and Methodology}
\label{sec:framework}
This section establishes a systematic framework to investigate the relationship between classification robustness and explanation robustness, ultimately developing our proposed solution (\ours) to address observed contradictions. We create a controlled level of explanation robustness using TRADES, then quantify explanation robustness through clustered evaluation, analyze loss landscape paradoxes, and finally introduce \ours for targeted landscape engineering.

\begin{figure}[t!]
\centering
\includegraphics[width=0.48\textwidth]{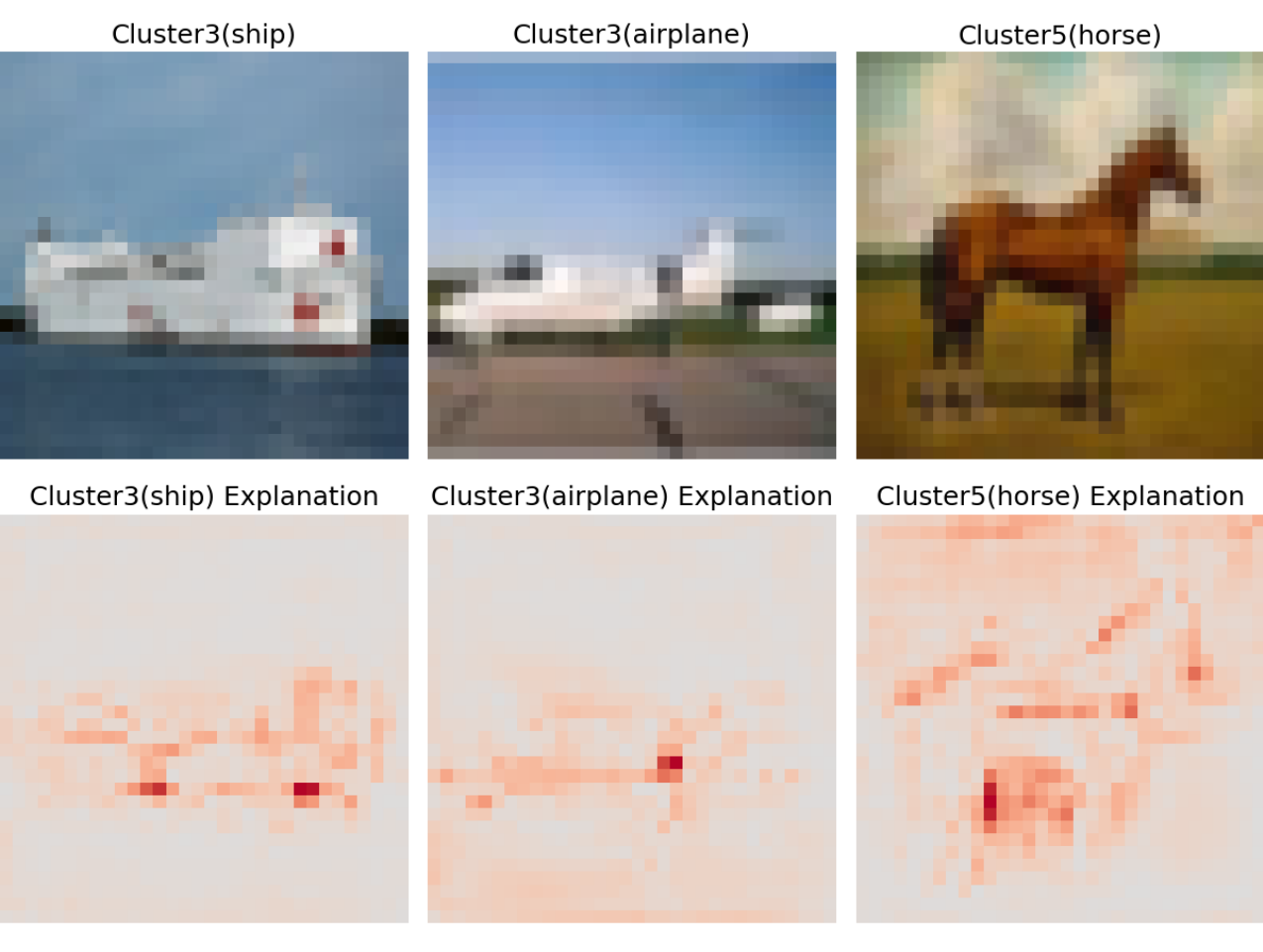}

\caption{Explanation consistency within clustered groups: (a) Cluster 3 exhibits focused attention on the shape of machines though they have different classification labels, while (b) Cluster 5 highlights the shape of horses.}
\label{fig:cluster_example}
\end{figure}

\subsection{Classification and Explanation Robustness}
\label{subsec:control}

\subsubsection{Classification Robustness}
Our systematic investigation begins by establishing precise control over model robustness characteristics through the TRADES framework~\citep{zhang2019theoretically}. TRADES provides fine-grained control of classification robustness via its parameterized loss function and the parameter $\alpha$:
\begin{equation}
\label{eq:trades}
\mathcal{L} _{\text{TRADES}} = \underbrace{\mathcal{L}_{\text{sc}}(f(x),y)}_{\text{Standard Classification Loss}} + \alpha \underbrace{\mathcal{L}_{\text{adv}}(f(x),f(x_{\text{adv}}))}_{\text{Adversarial Regularization}}
\end{equation}
where $f(x) \in \mathbb{R}^C$ denotes model outputs for input $x$ with $C$ classes, $x_{\text{adv}}$ is generated via projected gradient descent (PGD) attack~\cite{madry2017towards}: $x_{adv} = x + \epsilon_{adv}$ with $|\epsilon_{adv}|_\infty \leq \xi$, where the $\xi$ is a pre-defined constraint. The adversarial regularization term $\mathcal{L}_{\text{adv}}$ measures the KL divergence between original and adversarial predictions. The $\alpha$ parameter explicitly controls the tradeoff between clean accuracy and adversarial robustness, enabling controlled robustness levels. 

\vspace{3mm}
\noindent\textbf{Measuring Classification Robustness.} Following existing work~\cite{madry2017towards,zhang2019theoretically}, classification robustness is commonly measured by adversarial accuracy~($\accAdv$), which is the accuracy under adversarial attack on classification:
\begin{equation}
\accAdv := \frac{1}{N} \sum_{i=1}^{N} \mathbb{I} \left( f(x_{\text{adv},i}) = y_i \right),
\end{equation}
Where $\mathbb{I}$ is the indicator function. 
Compared with the clean accuracy ($\accClean := \frac{1}{N} \sum_{i=1}^{N} \mathbb{I} \left( f(x_{i}) = y_i \right)$) which measures classification performance, adversarial accuracy $\accAdv$ measures how good a model is against adversarial attacks on classification. Therefore, a higher $\accAdv$ indicates a better classification robustness.

\subsubsection{Explanation Robustness}
\label{subsec:eva}
Previous works on adversarial attacks have extended the framework to explanation through finding a perturbation $\delta^*$ with constrained optimization~\cite{tamam2022foiling}:
\begin{equation}
\delta^* = \argmin_{|\delta|_\infty \leq \xi} \lVert \expl(x_v+\delta) - \expl(x_t) \rVert_2
\end{equation}
where $x_v$ represents the victim (original) image, $x_t$ is the target image with desired explanation pattern, $f$ is the to-be-explained model, and $\expl(\cdot): \mathbb{R}^d \rightarrow \mathbb{R}^d$ denotes an explanation method (e.g., Grad-CAM~\cite{selvaraju2016grad} or Saliency Maps~\cite{simonyan2013deep}) that explains the behavior of model $f$ on an input image.

Explanation robustness is commonly measured by the following definition of attack loss: 

\begin{definition}[Explanation Attack Loss] Given a victim image $x_v$, a target image $x_t$ with desired explanation pattern, a to-be-explained model $f$ and an explanation method $\expl(\cdot)$, the explanation attack loss is
$\mathcal{L}_{e}(x_v,x_t) = \lVert \expl(x_v+\delta) - \expl(x_t) \rVert_2^2$, where $\delta$ is the optimization target.
\end{definition}

Since the ultimate goal of adversarial attacks on explanation is to manipulate the explanation of the victim image to resemble that of the target image, a higher explanation attack loss indicates a bigger difference between two explanations $\expl(x_v+\delta)$ and $\expl(x_t)$, which means a less successful attack.

\vspace{3mm}
\noindent\textbf{Generating $\mathcal{D}_e$ with cluster-based sampling.} In order to estimate the explanation robustness of a model $f$, existing methods use evaluation set $\mathcal{D}_e=\{(x_v, x_t)|x_v,x_t\in\mathcal{D}_{\text{origin}}$, where $\mathcal{D}_{\text{origin}}$ is the images in the original classification dataset. As the number of samples in $\mathcal{D}_{\text{origin}}$ increases, the size of $\mathcal{D}_e$ grows quadratically, which would be computationally infeasible. 

To ensure comprehensive coverage of the explanation space while maintaining computational traceability, we introduce a cluster-based sampling protocol grounded in explanation consistency.
Specifically, the protocol operates through three systematic steps:
\begin{enumerate}
\item \textbf{Feature Extraction}: Compute high-level representations using ResNet18's penultimate layer~\citep{he2016deep}, capturing semantically meaningful features for explanation consistency
\item \textbf{Clustering}: Apply clustering algorithm. In our experiment, we use K-means~\citep{lloyd1982least} with $k=10$ on CIFAR10. 
\item \textbf{Representative Sampling}: Select $N$ prototypes per cluster and forms evaluation set $\mathcal{D}_e$ for each cluster. In our experiment we set $N=15$ , forming evaluation set $\mathcal{D}_e$ with 150 images for 10 clusters and $150 \times 149 = 22,350$ unique $(x_v,x_t)$ pairs in total.
\end{enumerate}

As shown in Figure~\ref{fig:cluster_example}, images from the same cluster have similar explanations. We also report the explanation loss for intra-cluster and inter-cluster pairs to show that our clustering method indeed makes intra-cluster pairs share similar explanations quantitatively in \cref{tab:cluster_evaluation}, which shows intra-cluster pairs do have a smaller loss.


\begin{table}[H]
\centering
\begin{tabular}{lc}
\hline
ResNet18 & $\mathcal{L}_e^{\text{start}}$ ($\times 10^{-7}$)  \\
\hline
Intra-cluster  & 13.726 \\
Inter-cluster & 15.437 \\
\hline
\end{tabular}
\caption{Explanation loss at start of intra and inter clusters. The smaller explanation loss in the intra-cluster shows that images in the same cluster have similar explanations.}
\label{tab:cluster_evaluation}
\end{table}

\vspace{2mm}
\noindent\textbf{Measuring Explanation Robustness.} Specifically, we quantify the explanation robustness through the final explanation loss after explanation attack converges: $\mathcal{L}_e^{\text{end}} = \mathbb{E}_{(x_v,x_t)\sim\mathcal{D}_e}[ \mathcal{L}_e(x_v+\delta^{\text{end}}, x_t)]$, where $\delta^{\text{end}}$ is the optimal perturbation found by adversarial attacks on explanation. A higher $\mathcal{L}_e^{\text{end}}$ indicates that the attack struggles to manipulate the model’s explanations, suggesting stronger explanation robustness. 
We can also measure the explanation loss before the explanation attack starts: 
\begin{equation}
\begin{aligned}
\mathcal{L}_e^{\text{start}} = \mathbb{E}_{(x_v,x_t)\sim\mathcal{D}_e}[\mathcal{L}_e(x_v +\delta^{\text{start}}, x_t)],
\end{aligned}
\end{equation}
here $\delta^{\text{start}}$ is a random initial perturbation. It should be noted that $\mathcal{L}_e^{\text{start}}$ does not indicate the explanation robustness of a model since it is calculated before the explanation attack applies. However, providing $\mathcal{L}_e^{\text{start}}$ could let us know the difficulty of the attack before the beginning of the attack. Besides, we could also use the difference between $\mathcal{L}_e^{\text{start}}$ and $\mathcal{L}_e^{\text{end}}$ to roughly estimate the flatness of the loss landscape during training, while the definition for loss landscape at one point is introduced in the next section. After defining $\mathcal{L}_e^{\text{start}}$, we could obtain the quantitative results of explanation loss at start for the intra-cluster and inter-cluster for the clusters we obtain in the last section. In detail, \cref{tab:cluster_evaluation} represents the intra-cluster and inter-cluster results for $\mathcal{L}_e^{\text{start}}$ on ResNet18 and CIFAR10 dataset. From the results, we can see that $\mathcal{L}_e^{\text{start}}$ for intra-cluster is indeed smaller than inter-cluster results, indicating our cluster-based method can obtain similar explanations in the cluster.

\begin{table}[t!]
\centering
\caption{Comparison of classification robustness and explanation robustness of models trained with TRADES and different $\alpha$ on CIFAR10. Within a certain range, using the TRADES training method and increasing the value of $\alpha$ can not only improve the classification robustness but also improve the explanation robustness.}
\label{tab:trade_eva}

\begin{tabular}{@{}cccc@{}}
    \toprule
$\alpha$ & $\accAdv$ (\%) &  $\mathcal{L}_e^{\text{end}}$ ($\times10^{-7}$) & $\mathcal{L}_e^{\text{start}}$ ($\times10^{-7}$) \\ 
\midrule
0        & 0.00      & 6.206                        & 10.375                                           \\ 
0.5      & 23.57     & 10.640                       & 16.635                                           \\ 
1.0      & 28.31     & 10.946                       & 17.271                                           \\ 
2.0      & 31.77     & 10.965                       & 17.290                                           \\ 
4.0      & 33.28     & 11.293                       & 18.004                                           \\ 
5.0      & 33.98     & 11.469                       & 18.278                                           \\ 
10.0     & 34.87     & 11.592                       & 18.643                                           \\     \bottomrule
\end{tabular}
\end{table}

\subsubsection{Input Loss Landscape} An input loss landscape is a visualization of how a model's loss changes across different possible input values~\cite{alvarez2018robustness}, mapping the terrain of the loss function with respect to the input space, allowing researchers to understand how sensitive the model is to variations in the input data and identify potential robustness issues. Prior work has shown that a flatter input loss landscape for classification loss $\mathcal{L}_{\text{sc}}$ indicates stronger classification robustness~\cite{xie2020smooth,li2023understanding}. 

Following existing works~\cite{alvarez2018robustness}, we visualize the input loss landscape w.r.t explanation loss $\mathcal{L}_{e}(x_v,x_t;f)$ by plotting the change of $\mathcal{L}_{e}$ when adding a random noise $\mathbf{d}$ to the victim image $x_v$ with different magnitude~$\gamma$:

\begin{equation}
\label{eq:landscape}
    l(\gamma) = ||\expl(x_v + \gamma \mathbf{d})-\expl(x_t)||^2,
\end{equation}

\noindent where $\mathbf{d}$ is sampled from a standard Gaussian distribution. 


\subsection{Contradictory Findings on CIFAR10}
\label{subsec:finding}
Previous work has two assumptions: (1) a model with good classification robustness has a flat loss landscape w.r.t classification loss~\citep{xie2020smooth,li2023understanding}; (2) classification robustness and explanation robustness are positively correlated~\citep{boopathy2020proper,huang2023safari}. A natural conclusion would follow if the previous assumptions are true: a model with good classification robustness might have a good explanation and thus a flat loss landscape w.r.t explanation loss as well. 

\subsubsection{Step 1: Training models with different levels of classification robustness}

Our systematic investigation begins by establishing several models with different classification robustness through the TRADES framework~\citep{zhang2019theoretically} by training with different $\alpha$ in~\cref{eq:trades}.
The preliminary results on the CIFAR10 dataset~\cite{krizhevsky2009learning} can be found in~\cref{tab:trade_eva}. We can observe that with the increase of $\alpha$, the adversarial accuracy $\accAdv$ increases as well. This indicates that the model adversarially trained with TRADES shows better classification robustness with the increase of $\alpha$. This step provides us with models with different classification robustness.

\subsubsection{Step 2: Measuring explanation attack loss for models with different classification robustness} After getting the models with different classification robustness, the next step is to measure their explanation robustness, i.e., whether they perform differently under explanation attacks. With the assumption (2) classification robustness and explanation robustness are positively correlated~\citep{boopathy2020proper,huang2023safari}, we would expect that models with better classification robustness will show better explanation robustness. That is, models with higher $\accAdv$ will have higher $\mathcal{L}_e^{\text{end}}$. The preliminary results in~\cref{tab:trade_eva} shows that with the increase of $\alpha$, models indeed have higher $\mathcal{L}_e^{\text{end}}$. But since $\mathcal{L}_e^{\text{start}}$ is also higher with the increase of $\alpha$. Actually, calculating the $\Delta\mathcal{L} = \mathcal{L}_e^{\text{start}} - \mathcal{L}_e^{\text{end}}$, we could find that, after training, the attack is even more successful, considering the loss decrease. Therefore, we \textbf{cannot conclude} that better explanation robustness owes to better classification robustness. Next, we will explore a different way of measuring explanation robustness with the loss landscape. 

\begin{figure}[t!]
\centering
\includegraphics[width=0.85\linewidth]{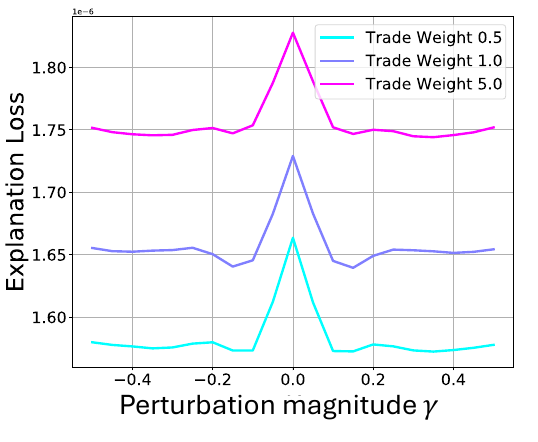}
\caption{Input loss landscape w.r.t explanation loss for models trained with different Trade Weight $\alpha$ in TRADES. The loss landscape does not show a clear difference between models that vary in explanation robustness because the loss change remains the same.}
\label{fig:flat_Trade}

\end{figure}

\subsubsection{Step 3: Connecting Robustness with Loss Landscape} To further investigate the explanation robustness of models, we visualize the input loss landscape w.r.t explanation loss. With the assumption (1) a model with good classification robustness has a flat loss landscape w.r.t classification loss~\citep{xie2020smooth,li2023understanding}, we would expect a similar analogy for explanation robustness:  a model with good explanation robustness has a flat loss landscape w.r.t explanation loss.

Using~\cref{eq:landscape}, we visualize the input loss landscape by plotting the change of explanation loss, and the results are shown in~\cref{fig:flat_Trade}. We also visualize the input loss landscape with normal training and Madry adversarial training (MAT) in Appendix~\cref{fig:flat_normal}. From~\cref{fig:flat_Trade}, we can find that the input loss landscape w.r.t. explanation loss \textbf{does not} show a difference in flatness, despite that their explanation loss $\mathcal{L}_e$ are different. Different from the conclusions drawn in classification robustness, models with good \ul{explanation robustness} do not exhibit a flat loss landscape w.r.t explanation loss.
This contradiction motivates us to further explore and propose a method to decouple classification and explanation robustness in the following section by changing explanation robustness by flattening the input loss landscape w.r.t explanation robustness, while maintaining classification robustness.

\subsection{Landscape-Aware Regularization}
\label{subsec:landcape}
Motivated by the observed landscape-robustness contradiction, in this section, we propose a method to decouple classification and explanation robustness. Specifically, we propose a new training algorithm to control the input loss landscape w.r.t explanation robustness and see if models with different flatness will perform differently on explanation robustness. If we can have a training algorithm that can influence explanation robustness while not changing classification robustness, we can conclude that classification robustness and explanation robustness are not strongly correlated.

To explicitly control the input loss landscape w.r.t explanation robustness, we propose \textbf{S}eparate \textbf{E}xplanation Robustness via \textbf{P}GD (\ours) through the following loss function:
\begin{equation}
    \mathcal{L}_{\ours} = \lVert I(x + \zeta) - I(x) \rVert^2_2,
\end{equation}
where $\zeta$ is a noise randomly sampled from a standard Gaussian distribution and $I$ is the explanation method. 
Note that off-the-shelf explanation AT~\citep{tamam2022foiling,dombrowski2019explanations} must be executed in a targeted setting: A victim image and a target image are required for the explanation of adversarial attacks. Calculating $\zeta$ through a targeted setting may increase the training time and increase the probability that the model overfits the chosen pairs. Therefore, we use randomly sampled noise, like for $\zeta$ which does not require a target image.

The new loss function $\mathcal{L}_{\ours}$ can be incorporated into existing training frameworks, including Madry adversarial training~\citep{madry2017towards}, TRADES~\citep{zhang2019theoretically}, and normal training. In this paper, we will mainly focus on Madry adversarial training plus the new training loss:
\begin{equation}
\label{eq:final_loss}
    \mathcal{L} = \mathcal{L}_{sc}(f(x_{adv}),y) + \lambda \mathcal{L}_{\ours}.
\end{equation}
where the hyperparameter $\lambda$ balances two components of the losses. When $\lambda>0$, we have $\ours_{\text{pos}}$ which guides the loss landscape to become flatter; when $\lambda<0$, we have $\ours_{\text{neg}}$, which guides the loss landscape to become sharper. The overall training algorithm is shown in~\cref{fig:SEP}. 

\begin{algorithm}[t!]
   \caption{\ours Algorithm}
   \label{fig:SEP}
   \small
\begin{algorithmic}[1]
   \STATE {\bfseries Input:} Dataset $\mathcal{D}$, total training iteration $T$, explanation method $I$, model weights $\mathbf{w}$, and balancing factor $\lambda$.
   \FOR{$t=0$ {\bfseries to} $T-1$}
   \FOR{batch $x$ in  $\mathcal{D}$}
   \STATE Sample a random noise $\zeta$ from a standard Gaussian distribution.
   \STATE Get adversarial samples (on classification): $x_{adv} = PGD(x,y)$.
   \STATE Calculate loss function with~\cref{eq:final_loss}.
   \STATE Update $\mathbf{w} \leftarrow \mathbf{w} - \eta \nabla \mathcal{L}(f(x),f(x_{adv}),y|\mathbf{w}) $
   \ENDFOR
    \ENDFOR
\end{algorithmic}
\end{algorithm}

In the following section, we show that our method can influence explanation robustness while it does not change classification robustness. Please note that our method is designed and used to explore the relationship between the classification robustness and explanation robustness instead of common dimensions that a normal algorithm will care like the performance. We also visualize the comparison of saliency maps from models trained with different algorithms to provide how our methods influence the saliency maps in~\cref{fig:saliency maps} in the later section to show how our method works.


%% file: 3experiment.tex
\section{Experiments}
\label{sec:experiment}
We conduct comprehensive experiments across multiple datasets and model architectures to validate our method's ability to decouple explanation robustness from classification robustness. Our evaluation addresses three key research questions:
\begin{itemize}
    \item \textbf{RQ1:} Can we independently control explanation robustness across different datasets and model architectures without changing classification robustness?
    \item \textbf{RQ2:} How do different explanation methods affect this relationship between two robustnesses?
    \item \textbf{RQ3:} Does our method generalize across different training protocols, e.g., different adversarial training methods like TRADES?
\end{itemize}

\begin{table*}[t!]
\caption{ Performance of models trained with ConvNet and ResNet18 on various datasets is evaluated using four training methods, w.r.t. $\mathcal{L}_e^{\text{end}}$ and $\accAdv$. Higher $\mathcal{L}_e^{\text{end}}$ indicates better explanation robustness; higher $\accAdv$ indicates better classification robustness. $\mathcal{L}_e^{\text{start}}$ is also included to show our method's influence on explanation robustness. The \textbf{best} performance in explanation and classification robustness and the \underline{worst} performance in explanation robustness are highlighted. There is no positive correlation between explanation and classification robustness achieved through $\ours_{pos}$ and $\ours_{neg}$ training methods, compared to MAT.
}
\scriptsize
\centering

\resizebox{\textwidth}{!}{
 \begin{tabular}{a|caca|caca}
    \hline
\rowcolor{Gray}
Model Architecture &\multicolumn{4}{c|}{ConvNet}& \multicolumn{4}{c}{ResNet18} \\
    \hline
\multicolumn{9}{c}{MNIST} \\
    \hline
Training Method& $\mathcal{L}_e^{\text{start}}$ ($\times 10^{-7}$)& $\mathcal{L}_e^{\text{end}}$ ($\times 10^{-7}$) & \accClean(\%)& \accAdv(\%)& $\mathcal{L}_e^{\text{start}}$ ($\times 10^{-7}$)& $\mathcal{L}_e^{\text{end}}$ ($\times 10^{-7}$)& \accClean(\%)& \accAdv (\%)\\
    \hline
Normal
&261.183 &204.825 &99.29 &0.00 &266.834 &146.16 &99.36 &0.00 \\
MAT&373.262 &298.729 &99.00 &89.92 &916.017 &778.003 &99.28 &\textbf{94.60} \\
$\ours_{pos}$& 93.033 & \underline{61.545} &98.8 &89.4 & 92.371 & \underline{59.278} &98.4 &91.63 \\
$\ours_{neg}$ & 806.204 & \textbf{657.180} &98.97 &\textbf{90.34} & 9356.306 &\textbf{8248.627} & 99.4 &93.95 \\
\hline
\multicolumn{9}{c}{FMNIST} \\
\hline
Training Method& $\mathcal{L}_e^{\text{start}}$ ($\times 10^{-7}$)& $\mathcal{L}_e^{\text{end}}$ ($\times 10^{-7}$) & \accClean(\%)& \accAdv(\%)& $\mathcal{L}_e^{\text{start}}$ ($\times 10^{-7}$)& $\mathcal{L}_e^{\text{end}}$ ($\times 10^{-7}$)& \accClean(\%)& \accAdv (\%)\\
\hline
Normal&106.530 &72.198 &92.32 &0.00 &128.640 &69.847 &91.57 &0.00 \\
MAT&386.370 &274.267 &62.85 &73.98 &588.610 &417.031 &79.22 &\textbf{67.10} \\
$\ours_{pos}$& 35.588 & \underline{22.465} &69.88 & \textbf{86.81} &32.466 & \underline{22.512} &68.75 &56.51 \\
$\ours_{neg}$&1811.969 & \textbf{994.818} &62.75 &76.89 &8050.942 &\textbf{7593.650} &70.23 &57.55 \\
\hline
\multicolumn{9}{c}{CIFAR10} \\
\hline
Training Method& $\mathcal{L}_e^{\text{start}}$ ($\times 10^{-7}$)& $\mathcal{L}_e^{\text{end}}$ ($\times 10^{-7}$) & \accClean(\%)& \accAdv(\%)& $\mathcal{L}_e^{\text{start}}$ ($\times 10^{-7}$)& $\mathcal{L}_e^{\text{end}}$ ($\times 10^{-7}$)& \accClean(\%)& \accAdv (\%)\\
\hline
Normal&10.375 &6.206 & 79.08 &0.00 &13.982 & \underline{6.130} & 81.32 &0.00 \\
MAT&16.913 &6.906 &64.85 &35.11 &31.959 &21.879 &67.22 &29.09 \\
$\ours_{pos}$& 3.565 & \underline{1.269} &64.94 & \textbf{35.25} & 11.962 &7.958 &66.68 & \textbf{29.69} \\
$\ours_{neg}$&19.002 & \textbf{7.590} &64.56 &34.86 &70.159 & \textbf{36.276} &39.17 &29.32 \\
\hline
\multicolumn{9}{c}{CIFAR100} \\
\hline
Training Method& $\mathcal{L}_e^{\text{start}}$ ($\times 10^{-7}$)& $\mathcal{L}_e^{\text{end}}$ ($\times 10^{-7}$) & \accClean(\%)& \accAdv(\%)& $\mathcal{L}_e^{\text{start}}$ ($\times 10^{-7}$)& $\mathcal{L}_e^{\text{end}}$ ($\times 10^{-7}$)& \accClean(\%)& \accAdv (\%)\\
\hline
Normal&10.099 & \underline{6.140} &48.39 &0.05 & 12.044 & \underline{4.716} &41.24 &0.00 \\
MAT&20.642 &13.650 &36.4 &17.35 &33.456 &22.623 &36.14 &15.70 \\
$\ours_{pos}$&13.650 &9.932 &37.41 & \textbf{17.98} &19.217 &12.744 &34.83 &15.16 \\
$\ours_{neg}$&22.506 &\textbf{14.970} &36.17 &17.43 & 35.525 & \textbf{24.289} &34.80 & \textbf{15.87} \\
\hline
\multicolumn{9}{c}{TinyImageNet} \\
\hline
Training Method& $\mathcal{L}_e^{\text{start}}$ ($\times 10^{-7}$)& $\mathcal{L}_e^{\text{end}}$ ($\times 10^{-7}$) & \accClean(\%)& \accAdv(\%)& $\mathcal{L}_e^{\text{start}}$ ($\times 10^{-7}$)& $\mathcal{L}_e^{\text{end}}$ ($\times 10^{-7}$)& \accClean(\%)& \accAdv (\%)\\
\hline
Normal& 0.966 & \underline{0.633} & 28.71 &0.00 &1.131 & \underline{0.528} &28.34 &0.00 \\
MAT&2.426 &1.728 &25.13 &9.55 &3.119 &2.349 &26.34 & 10.81 \\
$\ours_{pos}$&2.242 &1.571 &24.83 & \textbf{9.63} &1.967 &1.435 &25.96 & \textbf{10.83} \\
$\ours_{neg}$& 3.873 & \textbf{2.610} &24.31 &9.61 & 4.413 & \textbf{3.016} &26.11 &10.74 \\
\hline
\end{tabular}
}
\label{tab:main_conv}
\end{table*}

\subsection{Experimental Settings}

\subsubsection{Datasets} To thoroughly demonstrate the impact of our proposed training method and the resulting conclusions, we evaluate on five standard benchmarks: 
\begin{itemize}
\item [$\bullet$]  \textbf{CIFAR10} \cite{krizhevsky2009learning}: consisting of $60$k $32\times32$ color images in 10 classes including $50$k training images and $10$k test images.
\item [$\bullet$] \textbf{CIFAR100} \cite{krizhevsky2009learning}: containing the same images as CIFAR10 but has a more refined label with 100 categories, which makes it a harder dataset.
\item [$\bullet$] \textbf{MNIST} \cite{lecun1989backpropagation}: containing $60$k training samples and $10$k test samples from 10 digit classes. Each digit is a $28\times 28 $ grayscale image. 
\item [$\bullet$] \textbf{Fashion MNIST} \cite{xiao2017fashion}: consisting of $60$k training samples and $10k$ test samples from $10$ classes. Each sample is a $28\times 28 $ grayscale image in a clothes category. 
\item [$\bullet$] \textbf{TinyImageNet} \cite{le2015tiny}: it is a subset of ImageNet~\cite{deng2009imagenet} with 64x64 pixels and 200 categories
\end{itemize}

We also consider using ImageNet~\citep{deng2009imagenet} and the experiment results for ImageNet can be found in \cref{app:ImageNet}. The results for ImageNet are similar to the results here.

\subsubsection{Architecture of To-be-explained Model} 
In addition to utilizing diverse datasets, we have also designed four distinct model architectures for training on these datasets. We conduct experiments on ConvNet, ResNet~\citep{he2016deep}, Wide ResNet~\citep{zagoruyko2016wide} and MoblieNetV2~\citep{howard2017mobilenets,sandler2018mobilenetv2}. The ConvNet model consists of three convolutional layers and one fully connected layer from Gidaris et al.~\citep{gidaris2018dynamic}.  For ResNet and Wide ResNet, we use a standard ResNet18 and Wide-ResNet-28, respectively. We also adjust the ResNet, Wide ResNet, and MoblieNetV2 so that they can fit into all datasets we use. All models use Softplus activation for explanation attack compatibility~\citep{dombrowski2019explanations}, maintaining ReLU-like behavior with improved differentiability.

\subsubsection{Explanation Methods}
For explanation methods, we mainly use the implementations of five explanation methods from Captum~\citep{kokhlikyan2020captum}: Gradient~\citep{baehrens2010explain}, Gradient × Input~\citep{shrikumar2017learning}, Guided Backpropagation~\citep{springenberg2014striving}, Deep Lift~\citep{shrikumar2017learning} and Integrated Gradients~\citep{sundararajan2017axiomatic}. Their detailed descriptions can be found in \cref{app:data}.

\subsubsection{Training Protocols} We consider two baselines: normal training (Normal) and Madry adversarial training (MAT)~\citep{madry2017towards}.
We also explore two variants of the proposed method: $\ours_{pos}$ and $\ours_{neg}$, as mentioned in~\cref{subsec:landcape}. In the rest of this paper, unless specified, we will use $\lambda = 50000$ for $\ours_{pos}$ and $\lambda = -3000$ for $\ours_{neg}$. We use a learning rate of 0.01 for ConvNet and MobileNet while using a learning rate of 0.001 for ResNet and Wide ResNet.

\subsubsection{Hyperparameters}
For all experiments, we train our models for 25 epochs with 64 as the batch size. We also consider different training epochs and our conclusion remains the same as shown later.
To accelerate the training process, we use Adam~\citep{kingma2014adam} as the optimizer. We use the standard settings in adversarial training~\citep{pang2020bag}, with $\epsilon = 8 /255$ in PGD for RGB images and $\epsilon = 0.3$ for grayscale images, and steps in PGD are set to 10 for all experiments. We list the detailed hyperparameters in the Appendix~\cref{tab:main_hyperparameter}.

\subsubsection{Metrics}
As mentioned in~\cref{subsec:eva}, we measure explanation robustness using the explanation loss in the end after explanation attack $\mathcal{L}_e^{\text{end}}$. A higher $\mathcal{L}_e^{\text{end}}$ indicates a worse attack and thus better explanation robustness. We also report the explanation loss before explanation attack $\mathcal{L}_e^{\text{start}}$ to show the influence of our method on the explanation loss landscape. For classification robustness, we report adversarial accuracy $\accAdv$, with higher values indicating better classification robustness. Additionally, we include clean accuracy $\accClean$ to ensure the models function normally in non-adversarial settings.

\subsection{Decoupling Robustness Dimensions (RQ1)}
We extend our preliminary experiments on CIFAR10 in~\cref{subsec:finding} to multiple model architectures and datasets with Gradient × Input as the explanation method. The results are shown in
\cref{tab:main_conv} for ConvNet and ResNet18 with additional results for W-ResNet and MoblieNetV2 provided in the Appendix~\cref{tab:main_WRes}. 
We have the following observations:

\noindent $\bullet$ Across all the datasets and model architecture, $\ours_{pos}$ has the lowest $\mathcal{L}_e^{\text{end}}$, and $\ours_{neg}$ has the highest $\mathcal{L}_e^{\text{end}}$. This indicates that $\ours_{pos}$ shows weaker explanation robustness and $\ours_{neg}$ shows the strongest explanation robustness. The different performance w.r.t. explanation loss at end for $\ours_{pos}$ and $\ours_{neg}$ is mainly induced by the difference in $\mathcal{L}_e^{\text{start}}$, which is influenced by our training method by setting $\lambda$ to positive or negative.

\noindent $\bullet$ While $\ours_{pos}$, $\ours_{neg}$, and MAT have very similar $\accAdv$, $\ours_{pos}$ shows the weakest explanation robustness by having the lowest $\mathcal{L}_e^{\text{end}}$, and $\ours_{neg}$ shows the strongest explanation robustness. These results show that despite the classification robustness of $\ours_{pos}$, $\ours_{neg}$, and MAT being similar, their explanation robustness is different. Therefore, we argue that there is no inherent relationship between explanation robustness and classification robustness.

\noindent  $\bullet$ In the setting of CIFAR10 and ResNet18, increasing the explanation robustness by $\ours_{neg}$ hurts the $\accClean$ while it still does not change $\accAdv$, which represents classification robustness. This observation further validates our argument: classification robustness and explanation robustness may not be strongly correlated.

\noindent  $\bullet$  From the results, we could see that though $\ours_{neg}$ has a much higher $\mathcal{L}_e^{\text{end}}$ compared with the $\ours_{pos}$. However, if we consider $\Delta \mathcal{L} = \mathcal{L}_e^{\text{start}} - \mathcal{L}_e^{\text{end}}$, we can find  $\Delta \mathcal{L}$ for $\ours_{neg}$ is much higher than $\mathcal{L}_e^{\text{end}}$. This loss decrease demonstrates that our design is working since the larger loss decrease indicate $\ours_{neg}$ has a sharper loss landscape. Considering all the results, we hypothesis that the explanation robustness actually comes from the increase of the attack difficulty at the starting point instead of a flat loss landscape that is hard for the attacker to optimize. While previous works~\cite{xie2020smooth,li2023understanding} show that classification robustness comes from the flat loss landscape, this mechanism difference between classification and explanation robustness also further indicate that they might not be strongly correlated.

\begin{table*}[t]
\scriptsize
\caption{ 
Performance of different explanation methods (Gradient and Guide Propagation) in the training phase is evaluated w.r.t. $\mathcal{L}_e^{\text{start}}$, $\mathcal{L}_e^{\text{end}}$, $\accClean$ and $\accAdv$ on CIFAR10. Higher $\mathcal{L}_e^{\text{end}}$ indicates better explanation robustness, while higher $\accAdv$ denotes better classification robustness. The \textbf{best} and \underline{worst} performances in explanation robustness and classification robustness are highlighted. Under various explanation methods, $\ours_{pos}$ shows a lower explanation loss compared to $\ours_{neg}$, with similar adversarial accuracy.
}
\centering
\resizebox{\textwidth}{!}{
 \begin{tabular}{a|caca|caca}
\rowcolor{Gray}
\hline
Model Architecture &\multicolumn{4}{c|}{ConvNet}& \multicolumn{4}{c}{ResNet18} \\
\hline
\multicolumn{9}{c}{Gradient} \\
\hline
Training Method& $\mathcal{L}_e^{\text{start}}$ ($\times 10^{-7}$)& $\mathcal{L}_e^{\text{end}}$ ($\times 10^{-7}$) & \accClean(\%)& \accAdv(\%)& $\mathcal{L}_e^{\text{start}}$ ($\times 10^{-7}$)& $\mathcal{L}_e^{\text{end}}$ ($\times 10^{-7}$)& \accClean(\%)& \accAdv (\%)\\
    \hline
Normal&7.977 &4.591 &79.08 &0.00 & 11.310 & \underline{4.671} &81.32 &0.00 \\
MAT&13.810 &8.705 &64.85 & \textbf{35.11} &26.899 &18.215 &67.22 &29.09\\
$\ours_{pos}$& 0.876 & \underline{0.503} &52.89 &29.68 &11.317 &6.604 &66.76 & \textbf{37.69} \\
$\ours_{neg}$& 13.964 & \textbf{9.290} &53.23 &29.56 & 8282.990 & \textbf{7236.182} &49.38 &32.28 \\
\hline
\multicolumn{9}{c}{Guide Propagation} \\
\hline
Training Method& $\mathcal{L}_e^{\text{start}}$ ($\times 10^{-7}$)& $\mathcal{L}_e^{\text{end}}$ ($\times 10^{-7}$) & \accClean(\%)& \accAdv(\%)& $\mathcal{L}_e^{\text{start}}$ ($\times 10^{-7}$)& $\mathcal{L}_e^{\text{end}}$ ($\times 10^{-7}$)& \accClean(\%)& \accAdv (\%)\\
\hline
Normal&8.075 &4.639 &79.08 &0.00& 11.515 & \underline{4.736} &81.32 &0.00 \\
MAT&14.012 &8.813 &64.85 &35.11 &27.012 &18.311 &67.22 &29.09 \\
$\ours_{pos}$& 1.023 & \underline{0.506} &60.27 &33.57 &12.004 &7.593 &67.16 &30.64 \\
$\ours_{neg}$& 14.643 & \textbf{9.110} &59.74 & \textbf{33.78} & 27.422 & \textbf{18.940} &66.48 & \textbf{30.72} \\
\hline
\end{tabular}
}
\label{tab:diff_expl}
\end{table*}

\begin{figure}[t!]
     \centering
     \begin{subfigure}[b]{0.35\textwidth}
         \centering
         \includegraphics[width=\textwidth]{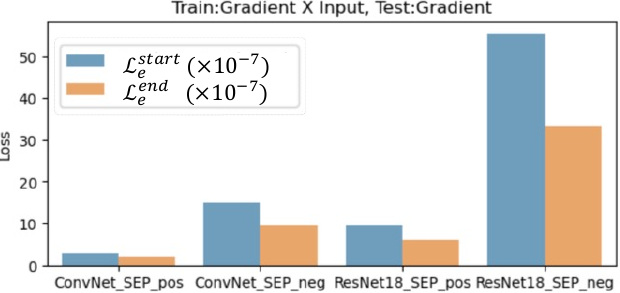}
     \end{subfigure}
     \begin{subfigure}[b]{0.35\textwidth}
         \centering
         \includegraphics[width=\textwidth]{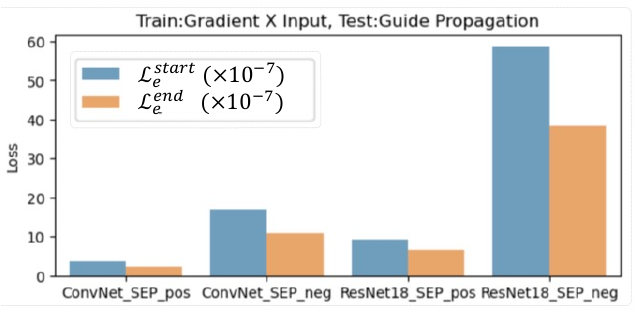}
     \end{subfigure}
\caption{Performance of varying explanation methods in the testing phase, w.r.t. $\mathcal{L}_e^{\text{start}}$, $\mathcal{L}_e^{\text{end}}$, $\accClean$ and $\accAdv$. Models are trained with Gradient x Input on CIFAR10 and tested on different explanation methods: Gradient (Top) and Guide Propagation (Bottom). Even if the explanation methods during training and testing are different, $\ours_{pos}$ shows a lower explanation loss compared to $\ours_{neg}$, while they have similar adversarial accuracy.}
\label{fig:transfer}

\end{figure}


\begin{figure}[th]
\centering
\includegraphics[width=\linewidth]{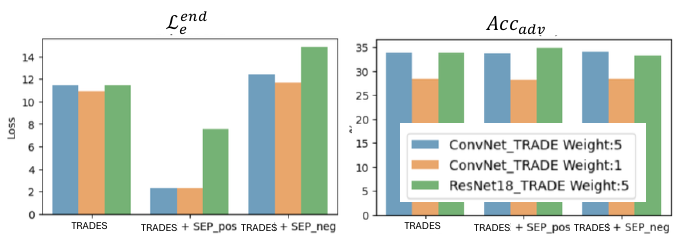}
\caption{The test results of the model trained using the TRADE training method with CIFAR10, combined with our approach, are presented. The findings indicate that when we apply our method to an alternative adversarial training method TRADES, distinct from MAT, we can still find that the classification robustness and explanation robustness are not inherently interconnected.}
\label{fig:Trade+our}
\end{figure}

\subsection{Explanation Method Agnosticism (RQ2)}

\subsubsection{Training Phase Variations} 

In the previous experiment, we demonstrated that under the Gradient × Input explanation method, our methods achieve similar classification robustness while exhibiting significantly different explanation robustness. To further investigate whether this conclusion holds for different explanation methods, we changed the explanation method to Gradient and Guide Propagation. The results are based on CIFAR10 and are summarized in~\cref{tab:diff_expl}. Extended results using DeepLIFT~\cite{shrikumar2017learning} and Integrated Gradients~\cite{sundararajan2017axiomatic} as the explanation methods on ConvNet and various datasets. The results using Guide Propagation with TinyImageNet and FMNIST can be found at \cref{tab:diff_expl_more}. We have the following observations:

\noindent  $\bullet$ Our methods achieve similar classification robustness under various explanation methods, yet they exhibit notably different explanation robustness. In most cases, $\ours_{pos}$ shows lower explanation loss compared to $\ours_{neg}$, despite similar adversarial accuracy.

\noindent  $\bullet$ Compared to MAT, our method $\ours_{pos}$ shows comparable adversarial accuracy, indicating similar classification robustness, but it demonstrates distinct explanation loss characteristics. This suggests that explanation robustness and classification robustness may not be strongly correlated.

\noindent $\bullet$ Comparing the results in \cref{tab:Deeplift}, \cref{tab:diff_expl} and \cref{tab:diff_expl_more}, we can find that no matter which explanation method we use or which dataset we use, we consistently get the conclusion that we could get quite different explanation robustness when we have the similar classification robustness for $\ours_{pos}$ and $\ours_{neg}$. This demonstrates the robustness of our results and demonstrates that our findings are not restricted to one specific explanation method or dataset.


\subsubsection{Testing Phase Generalization}
To test if our findings hold when using different explanation methods during testing, in this experiment, we use the same model trained with Gradient × Input (thus the classification robustness is the same for different testing phases), but change two different explanation methods (Gradient and Guide Propagation) in the testing phase.  
The results on CIFAR10 are shown in~\cref{fig:transfer}, where the detailed value of this experiment can be found in Appendix~\cref{tab:transfer}. While with the same classification robustness (as shown in~\cref{tab:main_conv}, under adversarial accuracy in CIFAR10), there is a huge difference between $\ours_{pos}$ and $\ours_{neg}$ w.r.t the explanation losses (both at the start and the end). This indicates that even with different explanation methods in the testing phase, the explanation robustness still does not show strong correlations with adversarial robustness.



\begin{figure*}[t!]
     \centering
     \begin{subfigure}[b]{0.32\textwidth}
         \centering
         \includegraphics[width=\textwidth]{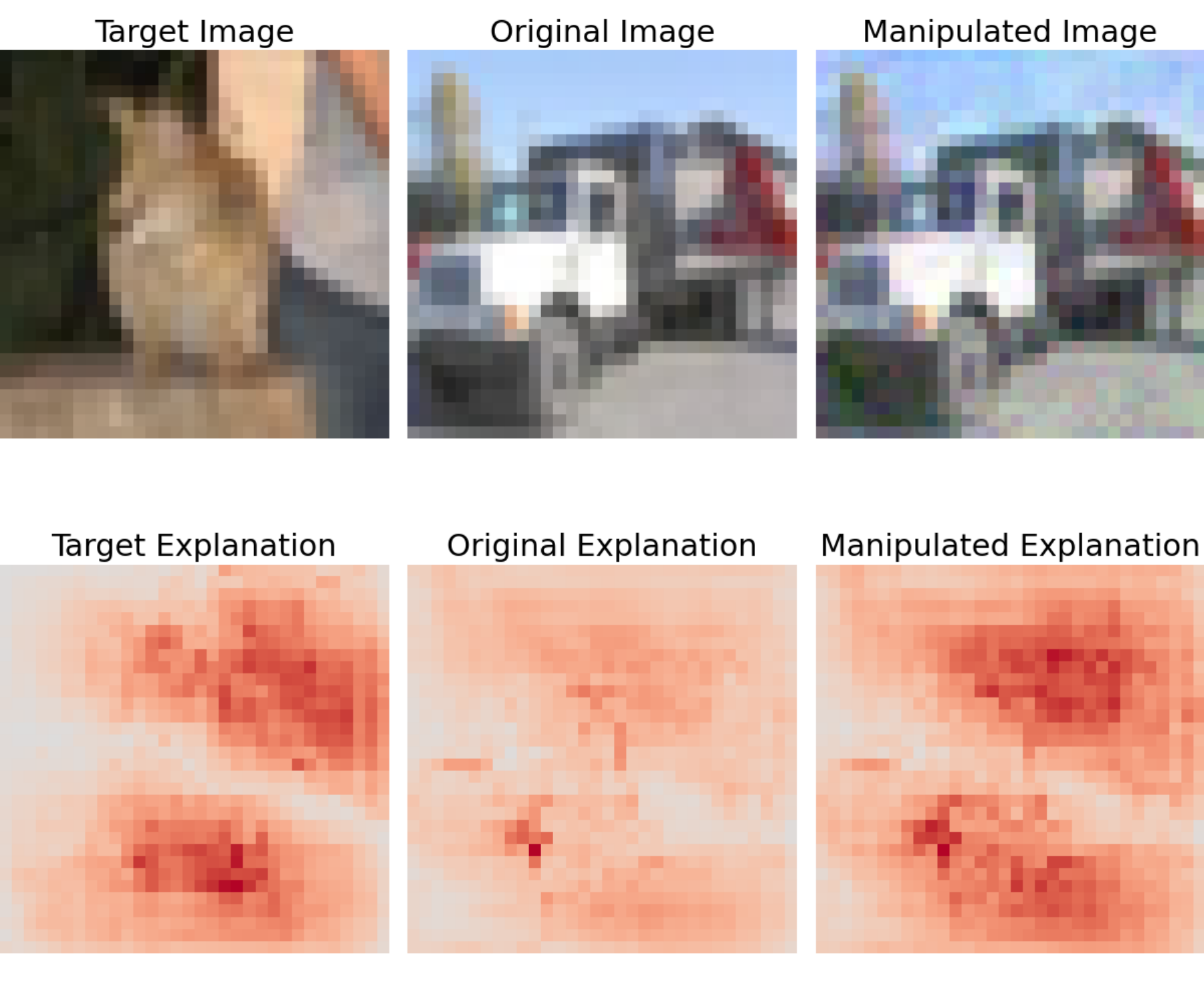}
         \caption{$\ours_{pos}$}
     \end{subfigure}
     \begin{subfigure}[b]{0.32\textwidth}
         \centering
         \includegraphics[width=\textwidth]{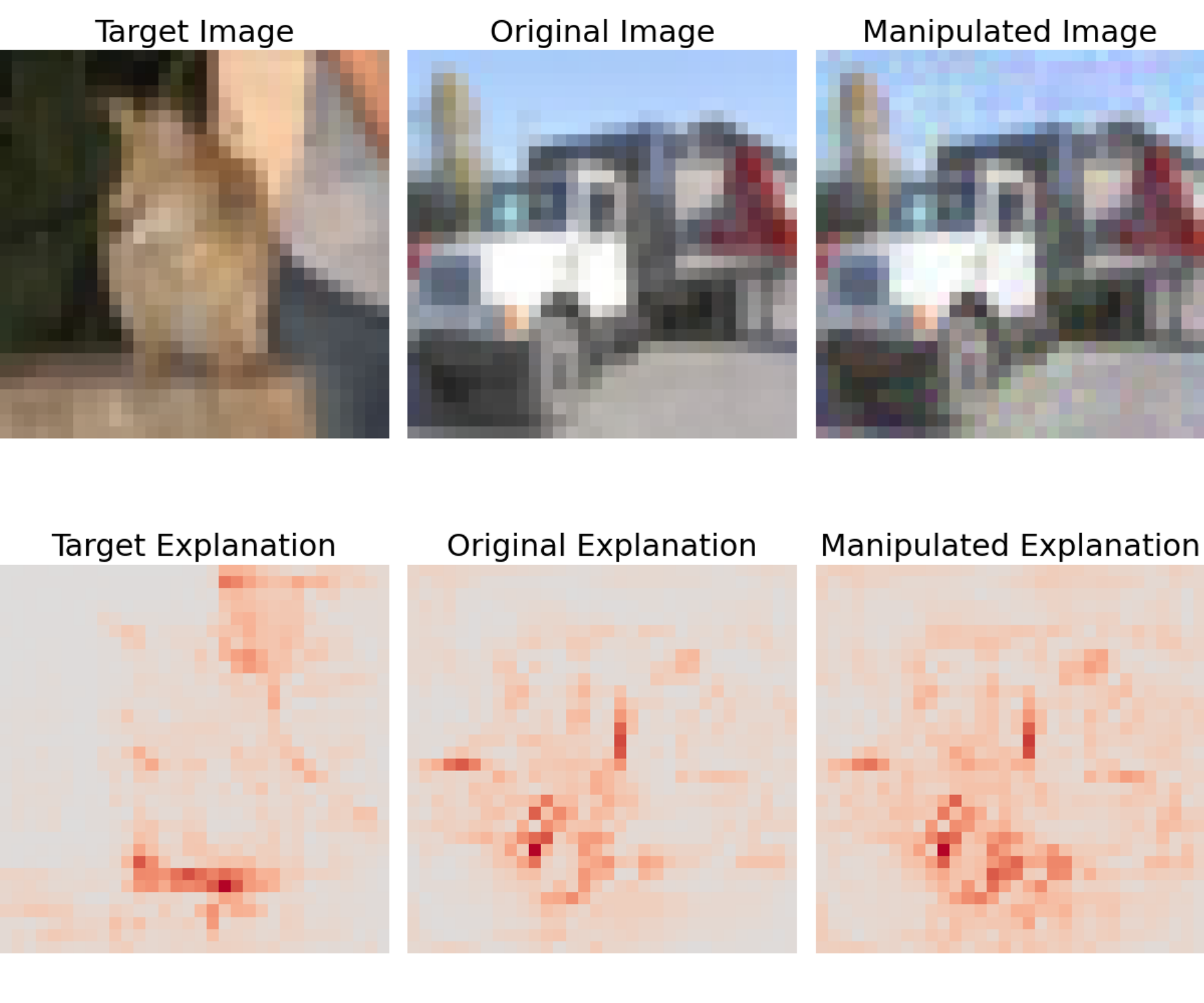}
         \caption{$\ours_{neg}$}
     \end{subfigure}
        \begin{subfigure}[b]{0.32\textwidth}
         \centering
         \includegraphics[width=\textwidth]{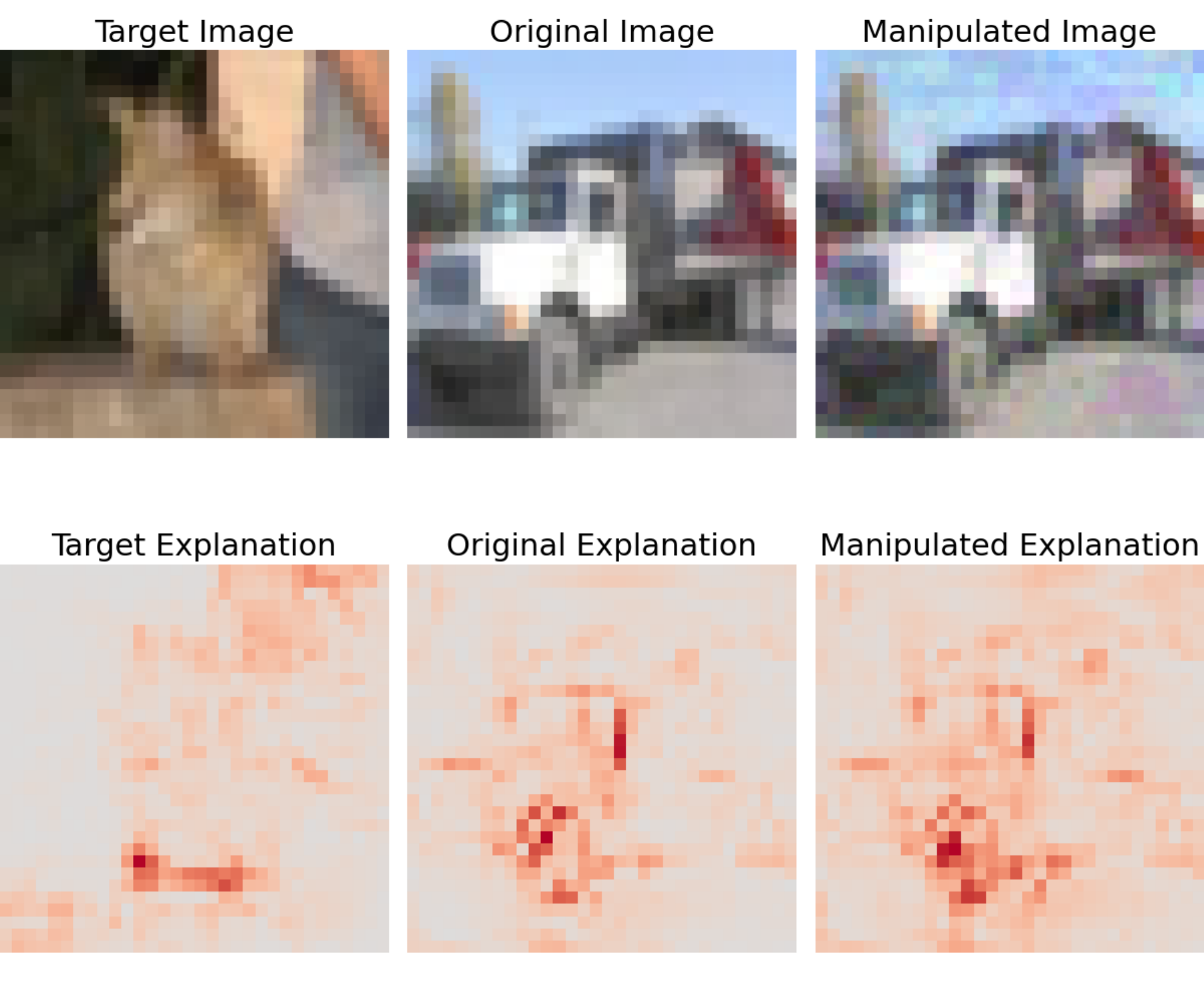}
         \caption{Adversarial Training}
     \end{subfigure}
\caption{Comparison of the saliency maps calculated from gradient x inputs on CIFAR10 with different training methods. Intuitively, $\ours_{pos}$ makes the model consider more input pixels, solely adversarial training makes the model consider only a few input pixels, while $\ours_{neg}$ considers even fewer input pixels compared with adversarial training. However, models trained with these three methods show the same classification robustness.}
\label{fig:saliency maps}
\vspace{-4mm}
\end{figure*}

\subsection{Training Protocol Generalization (RQ3)} 
All previous experiments utilized MAT~\citep{madry2017towards} as the default adversarial training protocol. To assess the generalizability of our approach across different adversarial training protocols, we employed TRADES~\citep{zhang2019theoretically} in this experiment and the results can be found in~\cref{fig:Trade+our} (details in Appendix~\cref{tab:trade_flat}). We can observe that when changing the adversarial training method from MAT to TRADES, a larger $\alpha$ will lead to a larger $\accAdv$, indicating a better classification robustness. As  $\alpha$ increases, $\mathcal{L}_e^{\text{end}}$ of the model trained with \ours varies.
This observation indicates that our $\ours$ method impacts explanation robustness without altering classification robustness, suggesting a weak correlation between explanation robustness and classification robustness.


\subsection{Parameter Sensitivity Analysis}
In this section, we examine how different parameters in our experiments affect the results. 

\noindent \textbf{Regularization Weights:} Firstly, we test how regularization weights $\lambda$ affect the results. In detail, we trained ConvNet networks on CIFAR10 with various $\lambda$ values. The test results are presented in ~\cref{tab:reg_weight}. We observe that the choice of $\lambda$ influences both the exploration rate at start and end. When $\lambda$ is greater than $10^4$ or less than $-3*10^3$, the explanation loss changes intensely.  However, from the results, we can see that for different $\lambda$ from $5 * 10^4$ to $-3 * 10^3$, the classification robustness remains the same, which further validate our hypothesis that explanation robustness and classification robustness may not be highly correlated.

\noindent\textbf{Training Epochs:} To demonstrate how training epochs might influence our conclusion, we conducted experiments on the ConvNet network using the CIFAR10 dataset with different training epochs. The results, as presented in \cref{tab:train_epoch}, indicate that the model's performance undergoes only marginal changes after 25 rounds for ConvNet, despite the epoch count continuing to increase. Therefore, we conclude that choosing 25 epochs in the rest experiments does not hurt the reliability of our argument. Besides, the results also support our conclusion. With the increase of training epochs, the classification robustness still increases while the explanation robustness actually decreases, which further validates our conclusion.

\begin{table}[h!]
\caption{ 
The evaluation of the ConvNet trained on CIFAR10 under different $\lambda$ conditions reveals that the relationship between explanation and classification robustness is not positively correlated when an appropriate $\lambda$ is selected during model training.
}
\centering
\resizebox{0.48\textwidth}{!}{%
\begin{tabular}{acaca}
\hline
$\lambda$ & $\mathcal{L}_e^{\text{start}}$ ($\times 10^{-7}$)& $\mathcal{L}_e^{\text{end}}$ ($\times 10^{-7}$) & \accClean(\%)& \accAdv(\%)\\ \hline
0~(MAT) &16.913  &6.206  &64.85  &35.11  \\
$5*10^4$ &3.565  &1.269  &64.94  &35.25  \\
$10^4$ &15.436  &5.870  &64.39  &35.18  \\
$10^1$ &17.646  &6.819  &64.45  &35.02  \\
$-10^2$ &17.820  &6.934  &64.67  &35.14  \\
$-3*10^3$ &19.002  &7.590  &64.56  &34.86  \\
\hline
\end{tabular}
}
\label{tab:reg_weight}
\end{table}

\begin{table}[t]
\scriptsize
\caption{
Test results of the ConvNet model at different training epochs on the CIFAR10 dataset. As the number of training epochs increases beyond 25, the improvement in performance is marginal. Therefore, 25 epochs are selected as the final number of training epochs for all models to maintain faster training speed without affecting the overall conclusions.
}
\centering
\resizebox{\columnwidth}{!}{
\begin{tabular}{a|c|c|c|c}
\rowcolor{Gray}
\hline
Model Architecture & \multicolumn{4}{c}{ConvNet (CIFAR10)} \\
\hline
Training Epoch & $\mathcal{L}_e^{\text{start}}$ ($\times10^{-7}$) & $\mathcal{L}_e^{\text{end}}$ ($\times10^{-7}$) & \accClean(\%) & \accAdv(\%) \\
\hline
25  & 4.388 & 1.605 & 64.94 & 35.25 \\
50  & 3.885 & 1.431 & 65.69 & 35.94 \\
75  & 3.671 & 1.378 & 66.33 & 36.27 \\
100 & 3.557 & 1.339 & 66.74 & 36.50 \\
\hline
\end{tabular}
}
\label{tab:train_epoch}
\end{table}

\begin{table*}[t]
\scriptsize
\caption{
Performance of using DeepLift and Integrated Gradients as explanation methods with ConvNet. Higher $\mathcal{L}_e^{\text{end}}$ indicates better explanation robustness, while higher $\accAdv$ denotes better classification robustness. The \textbf{best} and \underline{worst} performances in explanation robustness and classification robustness are highlighted. Under various explanation methods, $SEP_{pos}$ shows a lower explanation loss compared to $SEP_{neg}$, with similar adversarial accuracy.
}
\centering
\resizebox{\textwidth}{!}{
\begin{tabular}{a|caca|caca}
\rowcolor{Gray}
\hline
Explanation Method & \multicolumn{4}{c|}{DeepLift} & \multicolumn{4}{c}{Integrated Gradients} \\
\hline
\multicolumn{9}{c}{MNIST} \\
\hline
Training Method & $\mathcal{L}_e^{\text{start}}$ ($\times10^{-7}$) & $\mathcal{L}_e^{\text{end}}$ ($\times10^{-7}$) & \accClean(\%) & \accAdv(\%) & $\mathcal{L}_e^{\text{start}}$ ($\times10^{-7}$) & $\mathcal{L}_e^{\text{end}}$ ($\times10^{-7}$) & \accClean(\%) & \accAdv(\%) \\
\hline
MAT & 369.153 & 294.053 & 99.00 & 89.92 & 239.650 & 224.745 & 99.00 & 89.92 \\
$SEP_{pos}$ & 82.959 & \underline{57.408} & 98.93 & 95.97 & 76.284 & \underline{50.603} & 98.42 & \textbf{93.19} \\
$SEP_{neg}$ & 1101.038 & \textbf{896.157} & 98.97 & \textbf{96.16} & 778.663 & \textbf{534.113} & 98.68 & 92.76 \\
\hline
\multicolumn{9}{c}{FMNIST} \\
\hline
Training Method & $\mathcal{L}_e^{\text{start}}$ ($\times10^{-7}$) & $\mathcal{L}_e^{\text{end}}$ ($\times10^{-7}$) & \accClean(\%) & \accAdv(\%) & $\mathcal{L}_e^{\text{start}}$ ($\times10^{-7}$) & $\mathcal{L}_e^{\text{end}}$ ($\times10^{-7}$) & \accClean(\%) & \accAdv(\%) \\
\hline
MAT & 386.377 & 274.130 & 62.85 & \textbf{73.98} & 237.727 & 234.109 & 62.85 & \textbf{73.98} \\
$SEP_{pos}$ & 33.824 & \underline{21.429} & 60.75 & 65.52 & 21.425 & \underline{16.048} & 60.85 & 72.05 \\
$SEP_{neg}$ & 4739.769 & \textbf{3153.331} & 60.62 & 70.05 & 3624.231 & \textbf{2748.976} & 62.92 & 70.36 \\
\hline
\multicolumn{9}{c}{CIFAR10} \\
\hline
Training Method & $\mathcal{L}_e^{\text{start}}$ ($\times10^{-7}$) & $\mathcal{L}_e^{\text{end}}$ ($\times10^{-7}$) & \accClean(\%) & \accAdv(\%) & $\mathcal{L}_e^{\text{start}}$ ($\times10^{-7}$) & $\mathcal{L}_e^{\text{end}}$ ($\times10^{-7}$) & \accClean(\%) & \accAdv(\%) \\
\hline
MAT & 18.137 & 11.562 & 64.85 & 35.11 & 16.887 & 14.007 & 64.85 & \textbf{35.11} \\
$SEP_{pos}$ & 2.593 & \underline{1.273} & 65.76 & \textbf{35.38} & 3.696 & \underline{2.523} & 60.19 & 32.33 \\
$SEP_{neg}$ & 21.329 & \textbf{13.819} & 64.20 & 34.91 & 19.568 & \textbf{14.803} & 60.86 & 32.43 \\
\hline
\multicolumn{9}{c}{CIFAR100} \\
\hline
Training Method & $\mathcal{L}_e^{\text{start}}$ ($\times10^{-7}$) & $\mathcal{L}_e^{\text{end}}$ ($\times10^{-7}$) & \accClean(\%) & \accAdv(\%) & $\mathcal{L}_e^{\text{start}}$ ($\times10^{-7}$) & $\mathcal{L}_e^{\text{end}}$ ($\times10^{-7}$) & \accClean(\%) & \accAdv(\%) \\
\hline
MAT & 19.683 & 12.766 & 36.40 & 17.35 & 16.754 & 10.779 & 36.40 & \textbf{17.35} \\
$SEP_{pos}$ & 14.208 & \underline{9.201} & 39.10 & 18.23 & 5.320 & \underline{3.218} & 34.73 & 16.74 \\
$SEP_{neg}$ & 20.389 & \textbf{13.694} & 39.78 & \textbf{18.32} & 17.011 & \textbf{11.356} & 35.10 & 16.60 \\
\hline
\end{tabular}
}
\label{tab:Deeplift}
\end{table*}

\begin{table*}[t]
\scriptsize
\caption{
Performance of using Guide Propagation in the training phase with Fashion-MNIST and TinyImageNet. Higher $\mathcal{L}_e^{\text{end}}$ indicates better explanation robustness, while higher $\accAdv$ denotes better classification robustness. The \textbf{best} and \underline{worst} performances in explanation robustness and classification robustness are highlighted. Under various explanation methods, $SEP_{pos}$ shows a lower explanation loss compared to $SEP_{neg}$, with similar adversarial accuracy.
}
\vspace{-2mm}
\centering
\resizebox{\textwidth}{!}{
\begin{tabular}{a|caca|caca}
\rowcolor{Gray}
\hline
Model Architecture & \multicolumn{4}{c|}{ConvNet} & \multicolumn{4}{c}{ResNet18} \\
\hline
\multicolumn{9}{c}{Fashion-MNIST (FMNIST)} \\
\hline
Training Method & $\mathcal{L}_e^{\text{start}}$ ($\times10^{-7}$) & $\mathcal{L}_e^{\text{end}}$ ($\times10^{-7}$) & \accClean(\%) & \accAdv(\%) & $\mathcal{L}_e^{\text{start}}$ ($\times10^{-7}$) & $\mathcal{L}_e^{\text{end}}$ ($\times10^{-7}$) & \accClean(\%) & \accAdv(\%) \\
\hline
Normal & 30.932 & 18.451 & 92.79 & 0.00 & 66.131 & 29.110 & 91.57 & 0.00 \\
MAT & 97.726 & 72.402 & 62.85 & 73.98 & 608.486 & 467.815 & 79.22 & 67.10 \\
$SEP_{pos}$ & 48.368 & \underline{34.672} & 78.46 & 67.28 & 97.703 & \underline{78.354} & 77.55 & 62.09 \\
$SEP_{neg}$ & 542.540 & \textbf{425.948} & 65.07 & \textbf{77.17} & 4219.351 & \textbf{3839.408} & 80.11 & \textbf{72.21} \\
\hline
\multicolumn{9}{c}{TinyImageNet} \\
\hline
Training Method & $\mathcal{L}_e^{\text{start}}$ ($\times10^{-7}$) & $\mathcal{L}_e^{\text{end}}$ ($\times10^{-7}$) & \accClean(\%) & \accAdv(\%) & $\mathcal{L}_e^{\text{start}}$ ($\times10^{-7}$) & $\mathcal{L}_e^{\text{end}}$ ($\times10^{-7}$) & \accClean(\%) & \accAdv(\%) \\
\hline
Normal & 0.559 & 0.281 & 28.71 & 0.00 & 0.617 & 0.216 & 28.34 & 0.00 \\
MAT & 1.356 & 0.787 & 25.13 & 9.55 & 2.577 & 1.411 & 26.33 & 10.81 \\
$SEP_{pos}$ & 0.983 & \underline{0.625} & 25.16 & 5.97 & 1.767 & \underline{1.226} & 28.68 & 11.47 \\
$SEP_{neg}$ & 1.566 & \textbf{0.977} & 24.89 & 4.99 & 3.403 & \textbf{1.761} & 26.79 & 11.23 \\
\hline
\end{tabular}
}
\label{tab:diff_expl_more}
\vspace{-4mm}
\end{table*}

\subsection{Hessian Analysis}
\begin{figure}[th]
\centering
\includegraphics[width=\linewidth]{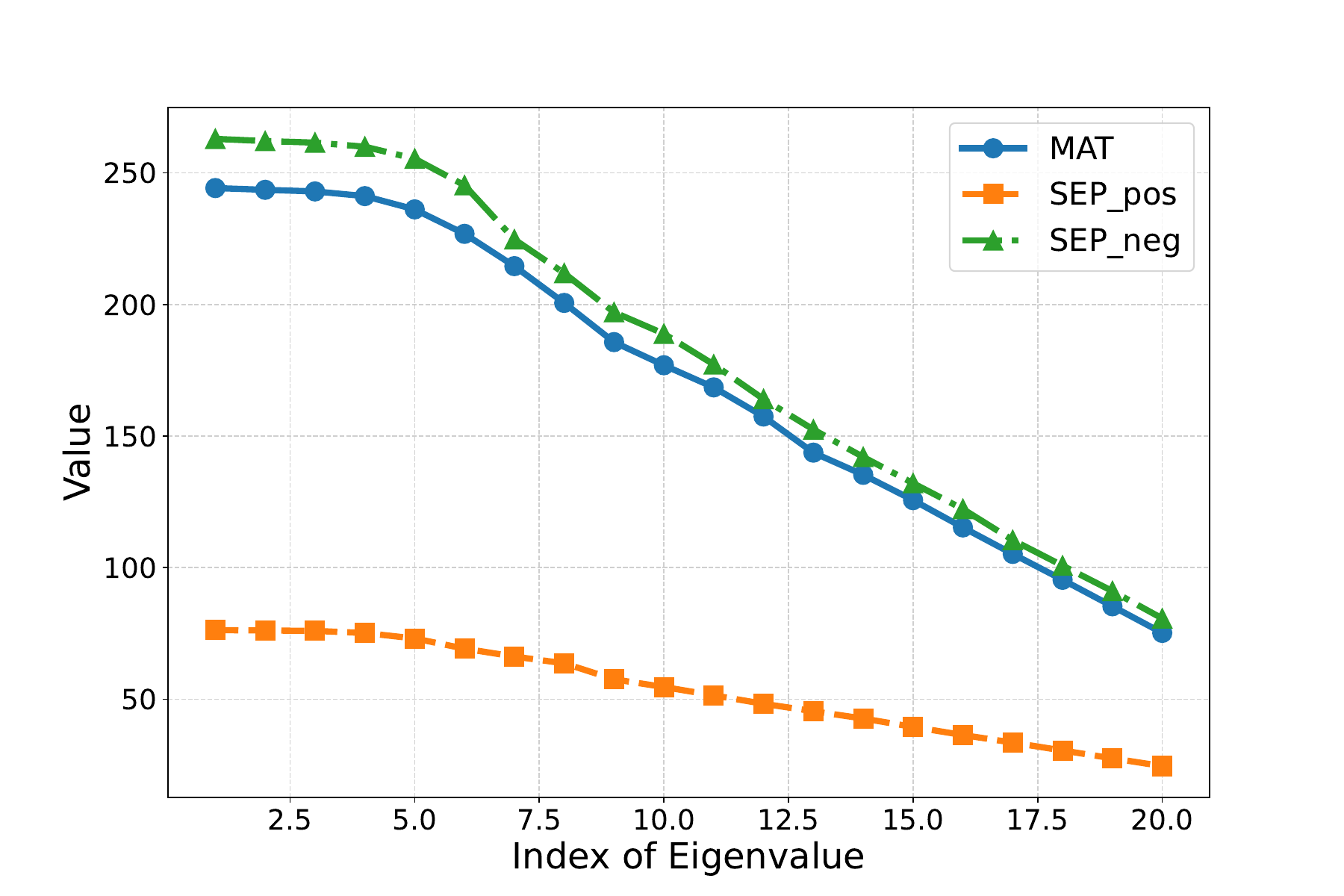}
\caption{Top-20 Eigenvalue of Hessian Matrix considering the explanation loss. A higher value of eigenvalue indicates a shaper loss landscape. The results show that \ours influence the loss landscape.}
\label{fig:hessian}
\vspace{-5mm}
\end{figure}

To further analyze how adversarial training and \ours affect the input loss landscape, we perform a Hessian analysis of the loss with respect to the input. Concretely, we study the input Hessian $H_x = \nabla_x^{2} L_e(x;\theta)$ 
 with $\theta$ fixed after training. The eigenvectors of $H_x$ define principal directions of curvature in input space and the corresponding eigenvalues quantify how rapidly the loss bends along those directions. Large absolute eigenvalues indicate a sharp loss landscape, while small absolute values indicate flatness of the loss landscape. Following previous works~\cite{liu2020loss}, we compute the top 20 eigenvalues at evaluation inputs and report their dataset means for the Hessian of explanation loss, which captures both how steep the sharpest direction is and whether sensitivity is concentrated in a few directions or spread across many, as reflected by the spectral decay. In detail, we show the results in \cref{fig:hessian}. The results show that $\ours_{pos}$ has much lower eigenvalues and thus indicates a much flatter loss landscape. Similarly, $\ours_{neg}$ has higher eigenvalues. These results firstly show that the design of \ours is successful because $\ours_{pos}$ and $\ours_{neg}$ influence the loss landscape as we want. Secondly, the results of flatness of loss landscape of $\ours_{pos}$ and $\ours_{neg}$ further validate the assumption in the previous section that the explanation robustness is mainly from the high initial loss instead of flat loss landscape, which indicates that there is no internal relationship between classification robustness and explanation robustness.

\subsection{Case Study}
In this section, we visualize the comparison of saliency maps from models trained with different algorithms to provide how our methods influence the saliency maps in~\cref{fig:saliency maps}. 
Specifically, we use the Gradient × Input method~\citep{shrikumar2017learning} to calculate saliency maps on the CIFAR-10 dataset~\citep{krizhevsky2009learning} with ConvNet.
From ~\cref{fig:saliency maps} we can see that while models trained with $\ours_{pos}$ exhibit broader activation regions in their saliency maps, models trained with standard adversarial training focus on fewer input pixels compared to $\ours_{pos}$. And models trained with $\ours_{neg}$ exhibit the narrowest focus, considering even fewer input pixels than adversarial training. Firstly, these findings highlight the flexibility of our method in shaping explanation patterns and suggest that explanation robustness can be independently controlled through targeted regularization techniques. Secondly, the results show that adjusting the loss landscape can lead to totally different explanation patterns. From the explanation patterns,  we can see that if the activated pixels are fewer, the initial distance, which is measured by the $\mathcal{L}_e^{\text{start}}$, will also be higher. Though in this case, the loss decrease is higher, the large initial distance ensures a better explanation robustness. On the other hand, if there are more activated pixels, it is more likely that two explanations are close at the beginning. Even though $\ours_{pos}$ has a flat loss landscape, which makes the attacker harder to optimize, the final attack is still successful. Based on the case, we further validate that the mechanism for explanation robustness is not based on the flat loss landscape, indicating there is no strongly correlated relationship between the two robustness.

%% file: 4conclusion.tex
\section{Conclusion}
This study challenges the widely held assumption that explanation robustness and classification robustness are strongly correlated. By systematically control classification robustness, we demonstrate that increasing explanation robustness does not necessarily result in a flatter input loss landscape for explanation loss. This finding contrasts with the well-established observation that enhancing classification robustness leads to a flatter input loss landscape for classification loss. These results reveal a fundamental difference in how these two types of robustness are influenced by adversarial training.
To address this discrepancy, we propose \ours, a novel algorithm that explicitly flattens the input loss landscape for explanation loss. Our experiments show that this approach effectively improves explanation robustness without affecting classification robustness, providing evidence that these two forms of robustness are not inherently linked. This decoupling highlights the importance of separately considering and optimizing explanation and classification robustness to ensure the reliability and trustworthiness of AI systems, particularly in high-stakes domains such as healthcare, autonomous driving, and finance.

Our findings emphasize the need for future research to further explore the relationship between these two types of robustness. In the future, we will theoretically investigate why adversarial training can improve explanation robustness in certain cases and what underlying mechanisms differentiate the behavior of classification and explanation robustness under adversarial attacks. A deeper understanding of these mechanisms will provide valuable insights into designing more robust and interpretable AI systems capable of maintaining both predictive accuracy and trustworthy explanations under adversarial conditions.

\section*{Acknowledgment}
The work was partially supported by NSF award \#2442477. We thank Amazon Research Awards, Cisco Research Awards, Google, and OpenAI for providing us with API credits. The views and conclusions in this paper are those of the authors and should not be interpreted as representing any funding agencies.

\newpage

%% file: 5appendix.tex
\newpage
\clearpage
\appendix

\begin{table*}[h!]
\scriptsize
\caption{Test results of models trained by Wide ResNet network and MobileNet network on various data sets according to four training methods. The results presented indicate that the performance of models trained using the Wide ResNet network and MobileNet network on different datasets suggests that there is no positive correlation between the model's explanation robustness and classification robustness achieved through the $SEP_{pos
}$ and $SEP_{neg}$ training methods, as compared to the MAT training method.}
\centering
\resizebox{\textwidth}{!}{
  \begin{tabular}{@{}l|c|c|c|c|c|c|c|c@{}}
    \toprule
\multicolumn{5}{c}{Wide ResNet}& \multicolumn{4}{c}{MobileNet} \\
  \midrule
\multicolumn{9}{c}{MNIST} \\
  \midrule
Method& $\mathcal{L}_e^{\text{start}}$ ($\times 10^{-7}$) & $\mathcal{L}_e^{\text{end}}$ ($\times 10^{-7}$)& \accClean (\%)& Adv Acc& $\mathcal{L}_e^{\text{start}}$ ($\times 10^{-7}$) & $\mathcal{L}_e^{\text{end}}$ ($\times 10^{-7}$)& \accAdv (\%) & \accAdv (\%) \\
  \midrule
Normal& 267.050 & 206.194 & 99.58 & 0.00 &287.061 & \underline{188.700} &99.08 &0.02 \\
MAT&842.648 &736.839 & 98.92 & \textbf{82.82} &4328.176 &3356.135 &98.29 &94.19 \\
SEP\_pos&109.383 & \underline{99.891} & 99.01 & 82.77 &319.629 & 273.256 & 98.36 &\textbf{94.25}\\
SEP\_neg&937.845 & \textbf{744.698} & 98.87& 82.71 &8134.157 & \textbf{4454.656} & 98.33 & 94.23\\
  \midrule
\multicolumn{9}{c}{FMNIST} \\
\hline
Method& $\mathcal{L}_e^{\text{start}}$ ($\times 10^{-7}$) & $\mathcal{L}_e^{\text{end}}$ ($\times 10^{-7}$)& \accClean (\%)& \accAdv (\%)& $\mathcal{L}_e^{\text{start}}$ ($\times 10^{-7}$) & $\mathcal{L}_e^{\text{end}}$ ($\times 10^{-7}$)& \accAdv (\%) & \accAdv (\%) \\
  \midrule
Normal& 120.037 & \underline{69.593} & 92.79&0.00 & 180.159 & \underline{103.941} &91.93 &0 \\
MAT&328.817 &257.523 & 78.10 & \textbf{68.26} &4470.448 &3571.210 &68.72 &57.19 \\
SEP\_pos&109.996 &74.324 & 77.69 & 67.79 &236.547 &172.200 & 65.11 & 57.42\\
SEP\_neg&398.006 & \textbf{304.927} & 78.21 & 68.05 &6032.190 & \textbf{4809.288} &66.86 & \textbf{58.16} \\
  \midrule
\multicolumn{9}{c}{CIFAR10} \\
  \midrule
Method& $\mathcal{L}_e^{\text{start}}$ ($\times 10^{-7}$) & $\mathcal{L}_e^{\text{end}}$ ($\times 10^{-7}$)& \accClean (\%)& \accAdv (\%)& $\mathcal{L}_e^{\text{start}}$ ($\times 10^{-7}$) & $\mathcal{L}_e^{\text{end}}$ ($\times 10^{-7}$)& \accAdv (\%) & \accAdv (\%) \\
  \midrule
Normal& 17.920 & \underline{8.029} & 85.47 & 0.16 & 14.797& \underline{6.551} &77.48 &0 \\
MAT&41.513 &27.136 & 60.01 & 24.22 &21.502 &13.223 &51.51 & \textbf{23.81} \\
SEP\_pos&26.343 &16.217 & 59.87& 24.89&14.756 &7.907 & 49.91 & 23.27 \\
SEP\_neg&43.278 & \textbf{27.575} & 60.15& 25.08 &26.811 & \textbf{16.420} & 35.43 & 15.30 \\
  \midrule
\multicolumn{9}{c}{CIFAR100} \\
  \midrule
Method& $\mathcal{L}_e^{\text{start}}$ ($\times 10^{-7}$) & $\mathcal{L}_e^{\text{end}}$ ($\times 10^{-7}$)& \accClean (\%)& \accAdv (\%)& $\mathcal{L}_e^{\text{start}}$ ($\times 10^{-7}$) & $\mathcal{L}_e^{\text{end}}$ ($\times 10^{-7}$)& \accAdv (\%) & \accAdv (\%) \\
  \midrule
Normal& 13.677 & \underline{5.606} & 59.13 & 0 &17.015 &9.351 &43.91 &0 \\
MAT&30.027 &18.389 & 36.69 & \textbf{16.12} &20.054 &10.836 & 21.19 & 8.64\\
SEP\_pos&22.046 &13.704 & 33.88 & 13.19 &15.234 & \underline{8.510} & 21.82 & \textbf{10.05} \\
SEP\_neg&31.889 & \textbf{20.045} & 35.74 & 15.55 &21.544 & \textbf{13.843} & 21.35 & 7.88\\
  \bottomrule
\end{tabular}
}
\label{tab:main_WRes}
\end{table*}

\section{Data}
\label{app:data}

We evaluate five explanation methods from Captum~\citep{kokhlikyan2020captum}:
\begin{itemize}
\item Gradient~\citep{baehrens2010explain}: Raw input gradients
\item Gradient×Input~\citep{shrikumar2017learning}: Element-wise product of inputs and gradients
\item Guided Backprop~\citep{springenberg2014striving}: Filtered gradient visualization
\item DeepLIFT~\citep{shrikumar2017learning}: Reference-based difference attribution
\item Integrated Gradients~\citep{sundararajan2017axiomatic}: Path integral of gradients
\end{itemize}

\section{More Visualization Results}
\begin{figure}[h]
     \centering
     \includegraphics[width=0.5\textwidth]{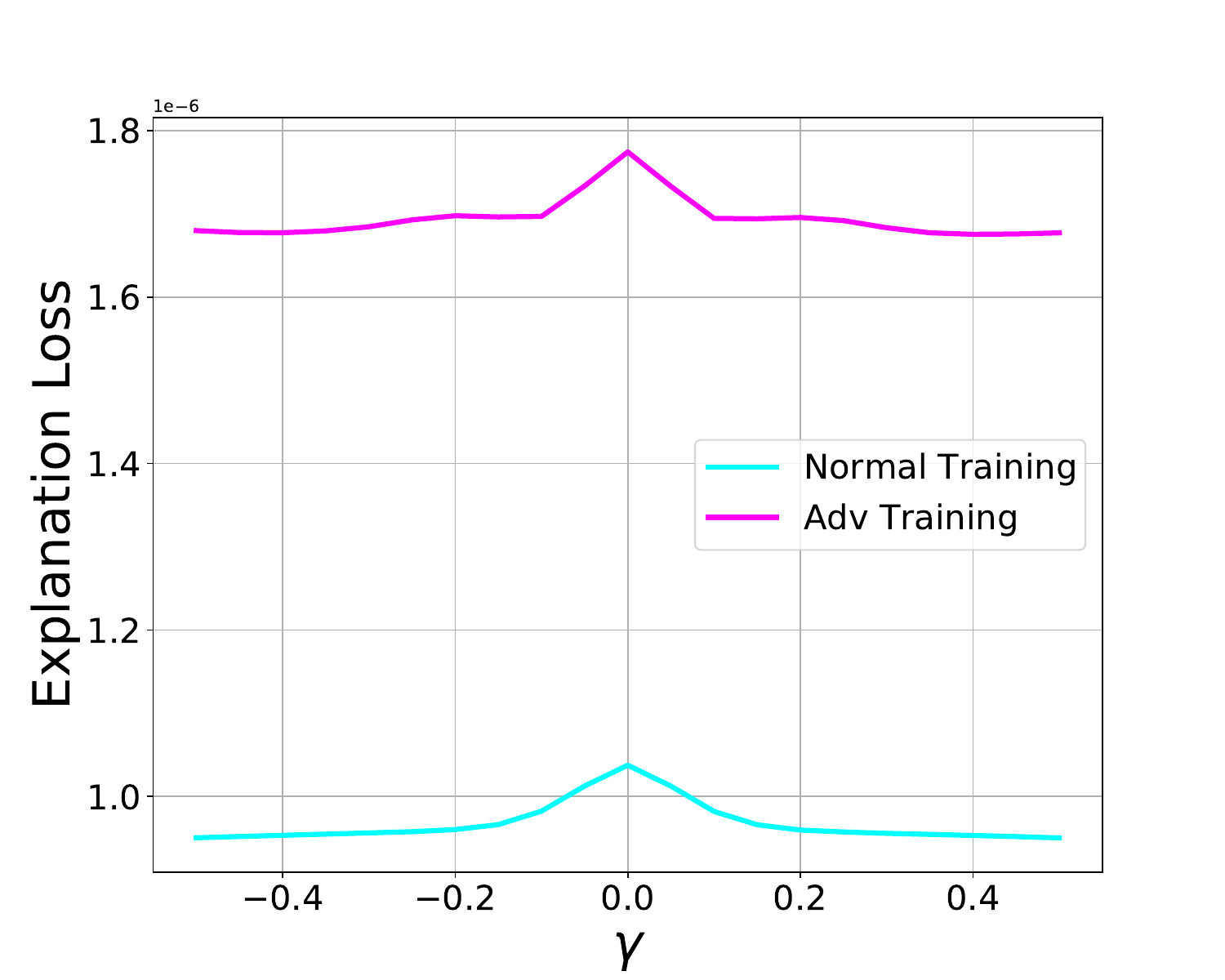}
     \caption{Comparison of input loss landscape w.r.t explanation loss with adversarial training and normal training. The results show that there is no obvious difference in input loss landscape.}
     \label{fig:flat_normal}
\vspace{-3mm}
\end{figure}

\begin{figure*}
 \centering
\begin{subfigure}{0.4\textwidth}
\includegraphics[width=\textwidth]{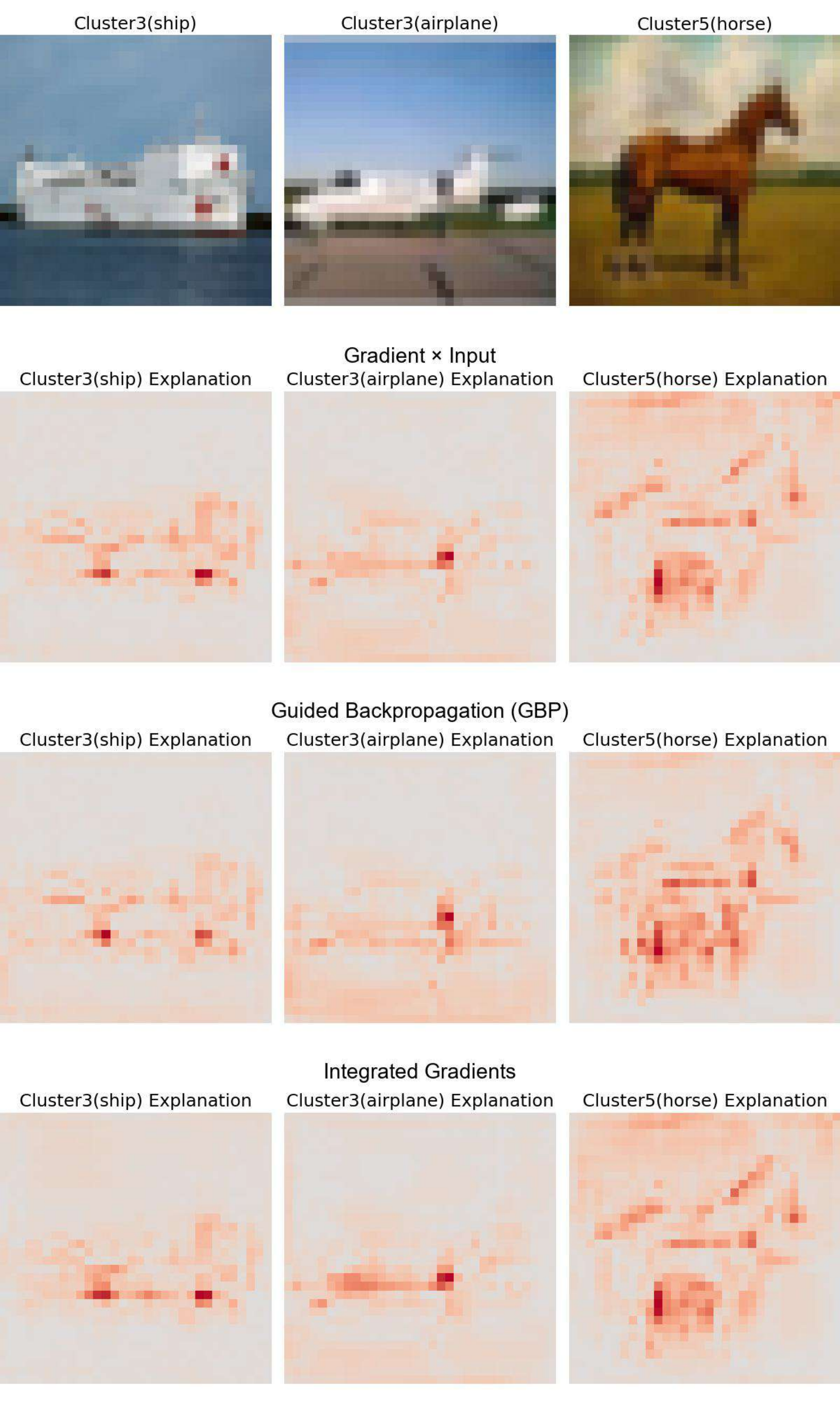}
\end{subfigure}
\begin{subfigure}{0.4\textwidth}
\includegraphics[width=\textwidth]{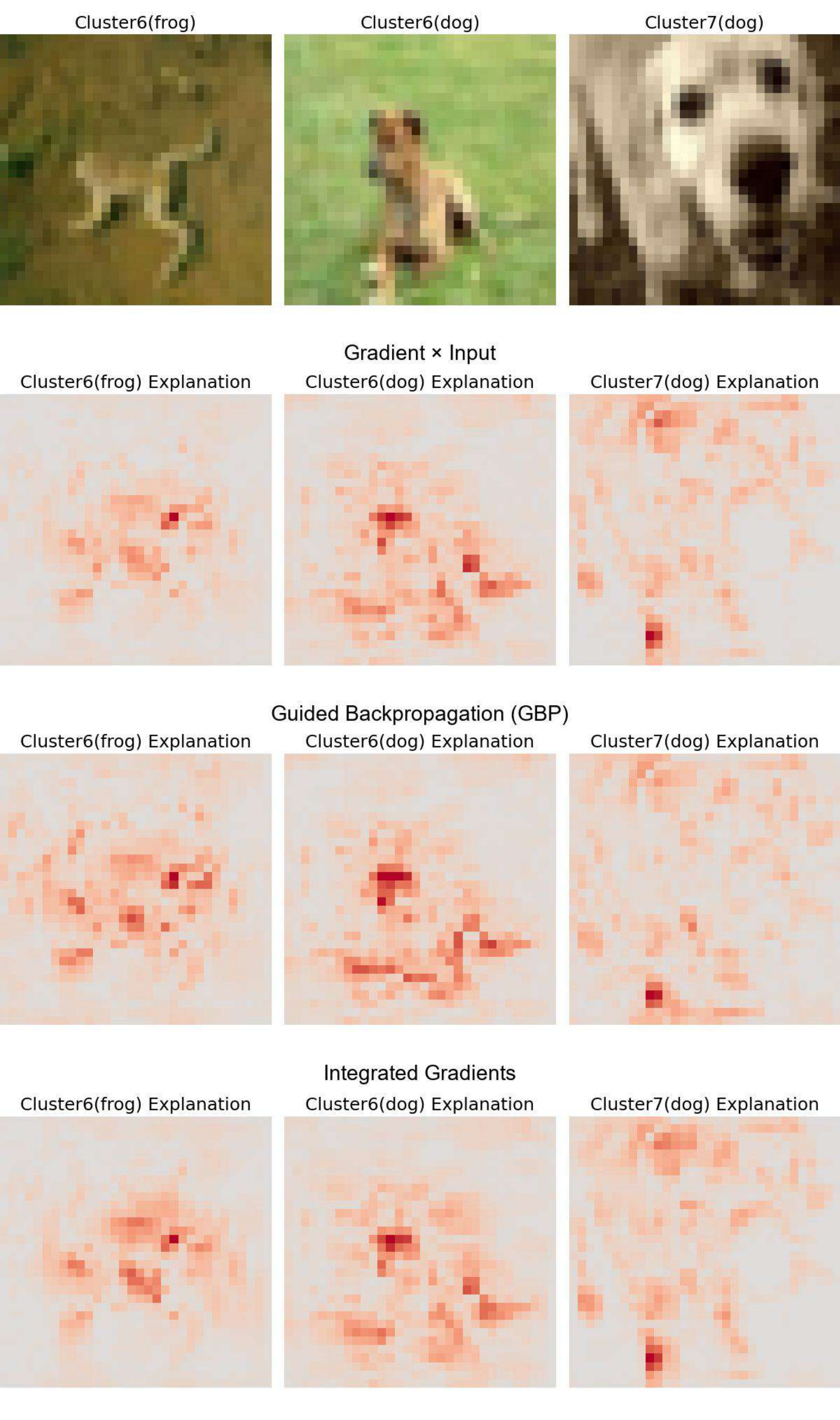}
\end{subfigure}
 \caption{Explanation of images from different clusters. The results show that images from the same cluster that even have different labels still have similar saliency maps on various explanation methods. Besides, images with the same label from different clusters still have different explanations. These results show that our method can sample the most representative subset of explanations.}
 \label{fig:cluster_combine}
\end{figure*}

Firstly, we visualize the input loss landscape w.r.t explanation loss using a normal trained model and model trained with Madry adversarial training in~\cref{fig:flat_normal}. The results show that increasing the explanation robustness does not flatten the input loss landscape.
Besides, we also visualize more saliency maps with more explanation methods with images from different clusters in \cref{fig:cluster_combine}. They all prove that we can choose the most representative saliency maps.

\section{Detailed Hyperparameter}

In this section, we provide the detailed hyperparameter for our CIFAR10 dataset in \cref{tab:main_hyperparameter}.

\begin{table}[H]
  \caption{Comparison of explanation loss for intra-cluster sample and inter-cluster sample on CIFAR10. The results show that our cluster method indeed the cluster images with similar explanations.
  }
  \label{tab:main_hyperparameter}
  \centering
  \begin{tabular}{@{}l|c|c@{}}
    \toprule
Models & Learning Rate  & $\lambda$   \\
     \midrule
\multicolumn{3}{c}{SEP\_pos} \\
    \midrule
ConvNet  & 0.01  & 5e4 \\
ResNet18 & 0.001 & 5e4 \\
Wide ResNet & 0.001 & 5e4 \\
MobileNet & 0.01 & 5e4\\
      \midrule
\multicolumn{3}{c}{SEP\_neg} \\
    \midrule
ConvNet  & 0.01 & -3e3 \\
ResNet18 & 0.001 & -1.9e3 \\
Wide ResNet & 0.001  & -1.9e3\\
MobileNet & 0.01 & -1.25e3 \\
  \bottomrule
  \end{tabular}
\end{table}

\section{More Experimental Results}
\label{app:exp}
\subsection{Experiments on W-ResNet and MobileNet}
\label{app:MobileNet}

We list the main results using Gradient X Inputs as training and testing explanation methods for W-ResNet and MobileNetV2 in \cref{tab:main_WRes}. We have the following observations:
\begin{itemize}
\item [$\bullet$] Once again, the adversarial accuracy for MAT, $SEP_{pos}$, and $SEP_{neg}$ is similar in most scenarios for W-ResNet and MobileNet, while $SEP_{pos}$ always has a smaller explanation loss compared with MAT, and $SEP_{neg}$ always has a larger explanation loss compared with MAT. These results show that influencing explanation robustness does not necessarily change classification robustness.

\item [$\bullet$] For W-ResNet and MobileNet, the adversarial accuracy for CIFAR100 fluctuates. For MobileNet and CIFAR100, compared with MAT, $SEP_{pos}$ increases classification robustness while $SEP_{neg}$ decreases it. However, this observation also indicates that the positive correlation between explanation robustness and classification robustness might not be true since $SEP_{pos}$ decreases explanation robustness while increasing classification robustness.

\end{itemize}

\subsection{Detailed Values for Transferablity Experiments}
The detailed values for Transferablity experiments can be found in \cref{tab:transfer} and the detailed values for experiments using TRADES for our method can be found in \cref{tab:trade_flat}. The analysis of these results can be found in the main paper.

\begin{table}[h!]
\scriptsize
\caption{Test results for transferability of explanation robustness. Models are trained with Gradient x Input and tested on different explanation methods.All models are trained on CIFAR10. Even if the interpretation methods during training and testing are different, comparing the training results of our proposed method with the AT training method of the corresponding configuration in \cref{tab:main_conv}, we can still draw our previous conclusions, which also shows that our conclusions are transferable.}
\centering
  \begin{tabular}{@{}l|c|c|c|c@{}}
\hline
\multicolumn{3}{c}{ConvNet}& \multicolumn{2}{c}{ResNet18} \\
\hline
\multicolumn{5}{c}{Train:Gradient X Input, Test:Gradient} \\
\hline
Method& $\mathcal{L}_e^{\text{start}}$ ($\times 10^{-7}$) & $\mathcal{L}_e^{\text{end}}$ ($\times 10^{-7}$)& $\mathcal{L}_e^{\text{start}}$ ($\times 10^{-7}$) & $\mathcal{L}_e^{\text{end}}$ ($\times 10^{-7}$) \\
\hline
$SEP_{pos}$&3.054 &1.901 &9.555 &5.903 \\
$SEP_{neg}$&15.093 &9.513 &55.526 &33.176 \\
\hline
\multicolumn{5}{c}{Train:Gradient X Input, Test:Integrated\_Grad} \\
\hline
Method& $\mathcal{L}_e^{\text{start}}$ ($\times 10^{-7}$) & $\mathcal{L}_e^{\text{end}}$ ($\times 10^{-7}$)& $\mathcal{L}_e^{\text{start}}$ ($\times 10^{-7}$) & $\mathcal{L}_e^{\text{end}}$ ($\times 10^{-7}$) \\
\hline
$SEP_{pos}$&3.767 &2.404 &9.209 &6.720 \\
$SEP_{neg}$&17.066 &10.923 &58.730 &38.433 \\
\hline
\end{tabular}
\label{tab:transfer}
\vspace{-2mm}
\end{table}

\begin{table*}[h!]
\small
\caption{The test results of the model trained using the TRADE training method, combined with our approach. The findings indicate that when we apply our method to TRADE, an alternative adversarial training method distinct from MAT, we can still deduce that classification robustness and interpretation robustness are not inherently interconnected. }
\centering
  \begin{tabular}{@{}l|c|c|c|c@{}}
\hline
\multicolumn{5}{c}{ConvNet} \\
\hline
\multicolumn{5}{c}{CIFAR10, TRADE Weight:5} \\
\hline
Method & $\mathcal{L}_e^{\text{start}}$ ($\times 10^{-7}$)  & $\mathcal{L}_e^{\text{end}}$ ($\times 10^{-7}$) & \accClean (\%) & \accAdv (\%) \\
TRADE &18.278&11.470&64.5&33.98  \\
TRADE + SEP\_pos& 3.878 & 2.285 & 63.84 & 33.85 \\
TRADE + SEP\_neg& 19.781 & 12.424  & 64.37  & 34.07 \\
\hline
\multicolumn{5}{c}{ConvNet} \\
\hline
\multicolumn{5}{c}{CIFAR10, TRADE Weight:1} \\
\hline
Method & $\mathcal{L}_e^{\text{start}}$ ($\times 10^{-7}$)  & $\mathcal{L}_e^{\text{end}}$ ($\times 10^{-7}$) & \accClean (\%) & \accAdv (\%) \\
TRADE & 17.271&10.965&72.63&28.31  \\
TRADE + SEP\_pos& 4.089 & 2.296 &72.41 & 28.20 \\
TRADE + SEP\_neg& 18.504 & 11.662 & 72.90 & 28.34 \\
\hline
\multicolumn{5}{c}{ResNet18} \\
\hline
\multicolumn{5}{c}{CIFAR10, TRADE Weight:5} \\
\hline
Method & $\mathcal{L}_e^{\text{start}}$ ($\times 10^{-7}$)  & $\mathcal{L}_e^{\text{end}}$ ($\times 10^{-7}$) & \accClean (\%) & \accAdv (\%) \\
TRADE & 18.278&11.469&64.50&33.98  \\
TRADE + SEP\_pos& 12.232&7.527&63.49&34.93  \\
TRADE + SEP\_neg& 22.571&14.881& 63.42 & 33.30 \\
\hline
\end{tabular}
\label{tab:trade_flat}
\vskip 0.5in
\end{table*}



\subsection{Experiments on ImageNet}
\label{app:ImageNet}

Here, we present our experiments on ImageNet with ResNet18 in \cref{tab:ImageNet}. We can find that the conclusion of ImageNet experiments is the same as the main paper: Increasing or decreasing explanation robustness will not necessarily influence the classification robustness.

\begin{table}[H]\scriptsize
\centering
\begin{tabular}{lcccc}
\hline
 & $\mathcal{L}_e^{\text{start}}$ ($\times 10^{-7}$) & $\mathcal{L}_e^{\text{end}}$ ($\times 10^{-7}$) & \accAdv~(\%) \\
\hline
Normal &114.70 & 63.52& 0.00\\
AT & 1281.71 &742.43 & 19.36  \\
$SEP_{pos}$ & 287.64 & 156.16 & 17.63 \\
$SEP_{neg}$ &  1427.33 & 905.25 & 17.44  \\
\hline
\end{tabular}
\caption{Experiments for ImageNet on ResNet18. The results are aligned with the conclusion made in the main paper.}
\label{tab:ImageNet}
\end{table}